\documentclass[10pt,twocolumn,letterpaper]{article}
\usepackage[T1]{fontenc}
\usepackage[latin9]{inputenc}
\usepackage{color}
\usepackage{array}
\usepackage{verbatim}
\usepackage{float}
\usepackage{booktabs}
\usepackage{mathrsfs}
\usepackage{multirow}
\usepackage{amsmath}
\usepackage{amssymb}
\usepackage{graphicx}
\PassOptionsToPackage{normalem}{ulem}
\usepackage{ulem}

\makeatletter

\providecommand{\tabularnewline}{\\}
\floatstyle{ruled}
\newfloat{algorithm}{tbp}{loa}
\providecommand{\algorithmname}{Algorithm}
\floatname{algorithm}{\protect\algorithmname}

\usepackage[pagenumbers]{cvpr}
\usepackage{times}
\usepackage{epsfig}
\usepackage{graphicx}
\usepackage{amsmath}
\usepackage{amssymb}

\usepackage{algpseudocode}
\usepackage{algorithm}
\usepackage{xcolor}
\usepackage{microtype}
\usepackage{colortbl}
\usepackage{booktabs}
\usepackage{tabu}
\usepackage{amssymb}
\usepackage{pifont}

\definecolor{cvprblue}{rgb}{0.21,0.49,0.74}
\usepackage[pagebackref,breaklinks,colorlinks,citecolor=cvprblue]{hyperref}
\def\paperID{****} 
\def\confName{CVPR}
\def\confYear{2024}

\makeatother

\usepackage{listings}
\begin{document}
\title{\textbf{Deployment Prior Injection for Run-time Calibratable Object
Detection}}
\author{Mo Zhou$^{1,2}$ \quad Yiding Yang$^{2}$ \quad Haoxiang Li$^{2}$
\quad Vishal M. Patel$^{1}$ \quad Gang Hua$^{2}$ \newline\\
$^{1}$Johns Hopkins University \qquad $^{2}$Wormpex AI Research}
\maketitle
\begin{abstract}
With a strong alignment between the training and test distributions,
object relation as a context prior facilitates object detection. Yet,
it turns into a harmful but inevitable training set bias upon test
distributions that shift differently across space and time. Nevertheless,
the existing detectors cannot incorporate deployment context prior
during the test phase without parameter update. Such kind of capability
requires the model to explicitly learn disentangled representations
with respect to context prior. To achieve this, we introduce an additional
graph input to the detector, where the graph represents the deployment
context prior, and its edge values represent object relations. Then,
the detector behavior is trained to bound to the graph with a modified
training objective. As a result, during the test phase, any suitable
deployment context prior can be injected into the detector via graph
edits, hence calibrating, or ``re-biasing'' the detector towards
the given prior at run-time without parameter update. Even if the
deployment prior is unknown, the detector can self-calibrate using
deployment prior approximated using its own predictions. Comprehensive
experimental results on the COCO dataset, as well as cross-dataset
testing on the Objects365 dataset, demonstrate the effectiveness of
the run-time calibratable detector.

\end{abstract}

\section{Introduction}

\label{sec:1}

Object detection~\cite{rcnn,fastrcnn,fasterrcnn} aims to find and
locate visual object instances (e.g., person, animal, vehicle, etc.)
within digital images. It has a wide range of applications in areas
such as autopilot, agriculture, and retail.

\begin{figure}
\includegraphics[width=1.0\linewidth]{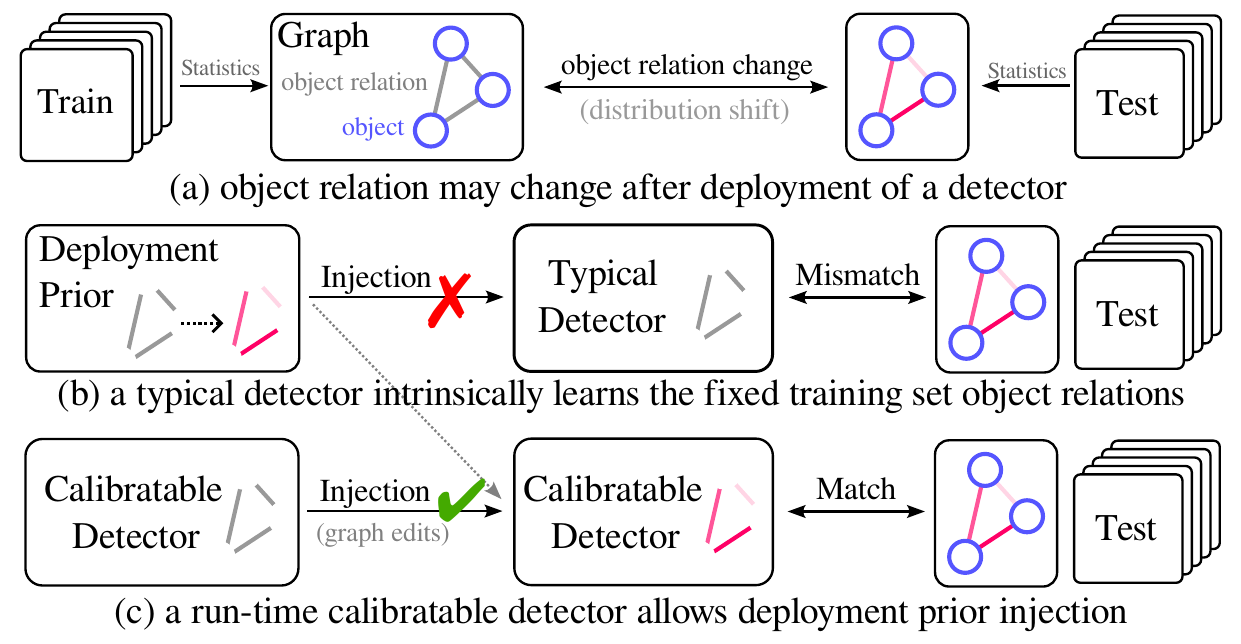}

\vspace{-0.3em}

\caption{Calibratable Object Detector that allows Deployment Prior Injection
at Run-time. The detector exposes a graph structure where the nodes
are objects, while the edges are object relations. The model behavior
is consistent with the graph structure, and hence deployment priors
can be injected as graph edits.}

\label{fig:calidet}

\vspace{-0.5em}
\end{figure}

Typically, the training data for a detector contains dense objects
of various kinds. Suppose the data distribution can be partly described
as a graph structure as shown in Fig.~\ref{fig:calidet} (a), where
the nodes represent object classes, and the edges represent the object
relations. In particular, we define ``object relation'' as the conditional
probability\footnote{Conditional probability among object classes is asymmetric. Its explicit
definition allows us to formulate expected model behavior intuitively.} that one object class appears given the presence of another object
class. 

Object relation can be used as a context prior to facilitate object
detection by assuming a match between the training and test data distributions.
However, in the case of a test data distribution shift, the already
learned contextual information from the training dataset will be regarded
as a harmful bias.

Upon deployment, object detectors often encounter such test distribution
changes due to variations in space (e.g., different geographical locations)
and time (e.g., different seasons). Such changes, however, can sometimes
be known in advance and regarded as ``deployment prior''. For instance,
Sprite may no longer co-occur with Coke after a shelf layout change
in a retail store; bikes often co-occur with a person on the road
but not in a shop. Expectedly, the detector will be influenced to
some extent by the object relations learned from the training dataset,
as demonstrated in Fig.~\ref{fig:calidet} (b). 

\begin{figure}
\centering

\includegraphics[width=0.47\columnwidth]{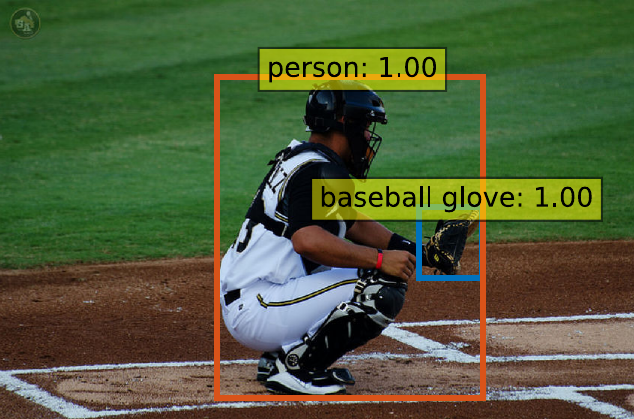}~~~~\includegraphics[width=0.47\columnwidth]{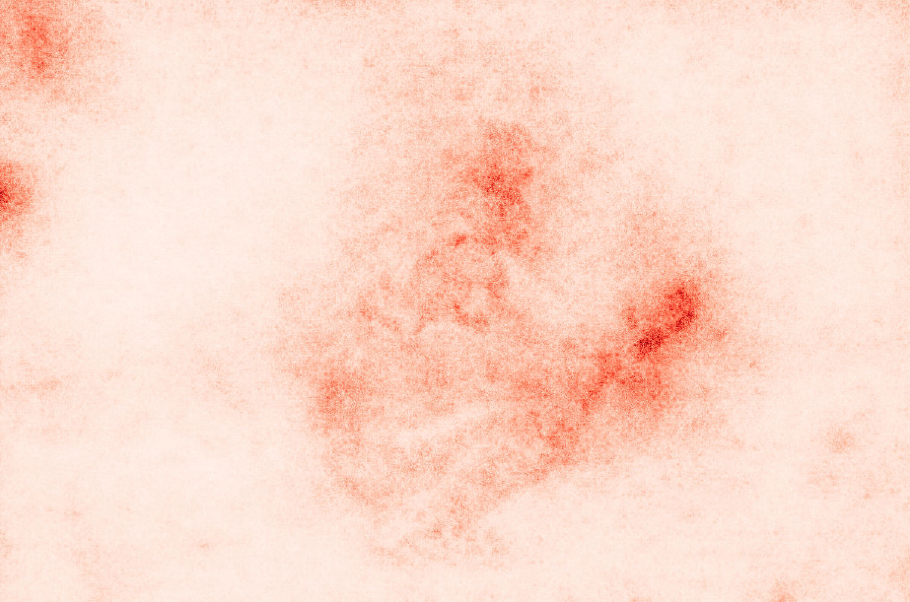}

\caption{A detector~\cite{detr} implicitly learns object relations via entangled
representations. Since $P(\text{person }|\text{ baseball glove})=99.0\%$
in the COCO training set, the gradient norm (shown on the right side)
of the ``baseball glove'' show the shape of ``person''. The gradient
norm is visualized following the attribution method in \cite{deformabledetr}.}

\label{fig:detr-demo}
\end{figure}

Take the COCO~\cite{coco} dataset as an example. There is a notable
relationship between the ``person'' and ``baseball glove'' classes,
because $P(\text{person }|\text{ baseball glove})=99.0\%$ in the
training dataset. As a result, a typical object detector, such as
DETR~\cite{detr}, implicitly learns the ``baseball glove'' representation
with ``person'' information entangled, as shown by the gradient
attribution of the ``baseball glove'' class in Fig.~\ref{fig:detr-demo}.
Such entanglement is a context prior if it remains in the test distribution,
or a bias if not. Nevertheless, it is permanently fixed once the training
is finished, and obscure to revise.

The existing works for relation modeling are statically designed to
either leverage~\cite{od-context1,od-relation1,od-relation2} or
mitigate (de-bias)~\cite{od-debias1,od-debias2,od-debias3} the impact
of context prior. But as long as the test distribution varies across
space and time, these methods will frequently encounter distribution
mismatch, and are not designed in a flexible way to adjust in run-time
to either leverage or mitigate the prior. As straightforward countermeasures,
fine-tuning or re-training with some new data upon every distribution
change is an inefficient and endless loop, due to the non-static nature
of the test distribution.

In contrast, we propose to learn a detector that can be calibrated
to align the relation distribution at run-time without any parameter
tuning, as shown in Fig.~\ref{fig:calidet} (c). Specifically, a
directed graph structure is exposed from the detector, where the nodes
are object class embeddings, and the edges are weighted by the conditional
probabilities (relations) among classes. The detector behavior is
trained to be bounded by the graph structure. Thus, deployment priors
can be injected into the detector as graph edits in edge weights,
where a large weight means to leverage the contextual cue from another
class, and a small weight means to mitigate such contextual cues.
In this way, the detector can be calibrated at run-time to better
match a shifted test data distribution.

{} A calibratable detector can be used as a drop-in replacement for
a general detector. Even if the deployment prior is not available,
the detector can be calibrated using its own prediction results through
``self-calibration''. Thus, through either manually injected prior
or self-calibration, the model can better match the shifted test distribution
with object relation changes without fine-tuning, let alone re-training.

{} We evaluate the calibratable detector using different deployment
priors. Notably, the more accurately the prior can describe the test
data distribution shift, the better it performs. In the self-calibration
scenario, the detector can improve object detection performance based
on deployment prior approximated using its own predictions.

\textbf{Contributions.} To the best of our knowledge, this is the
first work of a calibratable object detector, where deployment priors
can be injected at run-time to adjust the detector behavior for shifted
object relation distributions.

\section{Related Works}

\label{sec:2}
\begin{figure*}[t]
\includegraphics[width=1.0\linewidth]{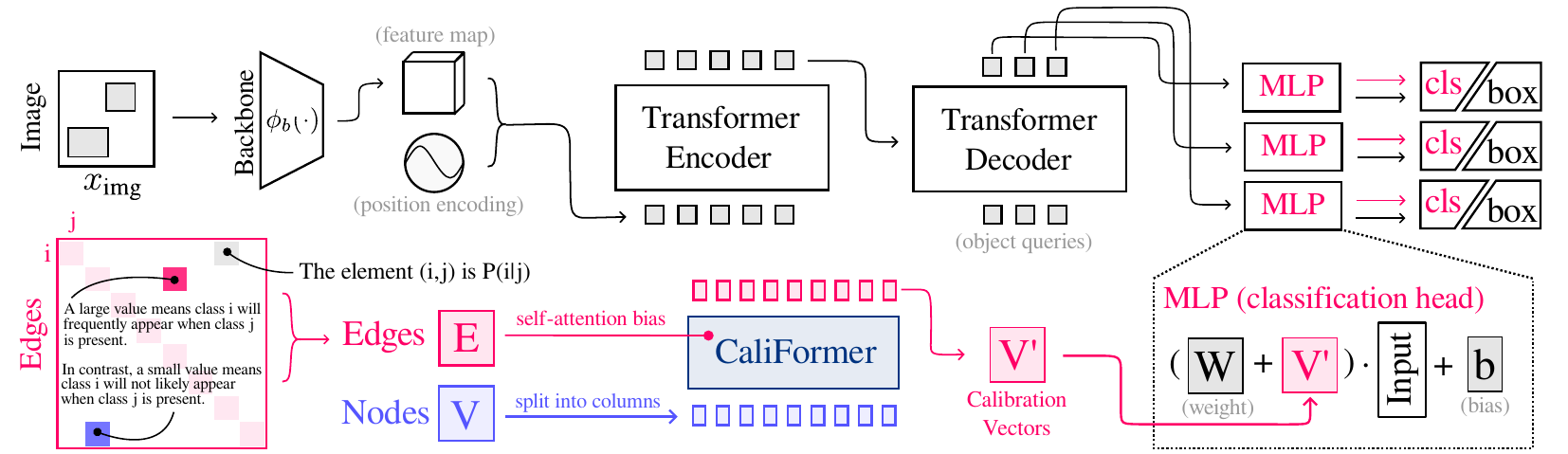}

\caption{The architecture of our proposed run-time calibratable object detector
(abbr., CaliDet). Injection of deployment prior is achieved by editing
the edges $E$ of the graph for the model. It can be used in DETR-like
detectors~\cite{deformabledetr,detr,dino}, such as DINO~\cite{dino}.
This figure shows the CaliFormer for calibration vectors $V'$ which
is elaborated in Section~\ref{sec:31}. The other components in our
model are elaborated in Section~\ref{sec:32}.}

\label{fig:architecture}
\end{figure*}
\textbf{Object Detection} methods can be categorized into three groups:
(1) two-stage detectors~\cite{rcnn,fastrcnn,fasterrcnn}; (2) one-stage
detectors~\cite{ssd,focalloss,yolo}, and (3) DETR family~\cite{detr,deformabledetr,dino,sam-detr,smca,sparse-detr,adamixer22cvpr,dn-detr}.
Specifically, DETR~\cite{detr} eliminates manually designed components
and employs the Transformer~\cite{transformer}. A deformable attention~\cite{deformabledetr}
is proposed to improve convergence speed and small object detection
performance. The existing detectors tend to implicitly learn permanently
fixed, data-specific, and non-calibratable object relations. Instead,
we aim to make a detector run-time calibratable to adapt to distribution
change without any parameter update.

\textbf{Relation Modeling} in object detection aims to build the relations
between the objects in the image and facilitate object detection.
Such context modeling is beneficial when the training and test distributions
are similar~\cite{relation_distill,spatial_aware_graph_relation,deep_regionlets,od-context1,od-relation1,od-relation2}.
Based on granularity, they can be divided into three groups: (1) feature-level~\cite{dynamic_head},
(2) part-level~\cite{relationnet++}, and (3) object-level~\cite{relation_network}.
For example, \cite{relation_network} uses the attention mechanism
to build the relationship among detected proposals and refine their
features. When the distributions mismatch, the relations learned by
the model are turned into a harmful bias~\cite{od-debias1,od-debias2,od-debias3}.
In contrast to modeling the relationship implicitly by the model,
we build them explicitly from a global view. Specifically, a graph
with editable edges is exposed to specify whether to leverage or mitigate
these object relations, making the detector calibratable at run-time.

\textbf{Meta Learning }is widely explored in computer vision~\cite{meta_rareobjects,meta_detr_fewshot,meta_incremental,maml,meta_sr}.
The goal of meta-learning is to train a model on a variety of tasks,
such that it can better adapt to new tasks~\cite{maml}. If we regard
different deployment priors as ``tasks'', the training of a calibratable
detector aims to generalize against different priors by predicting
a part of the model parameters, which will be detailed later.

\textbf{Visual Prompting}~\cite{exploring-visual-prompts,visual-prompt-tuning}
aims to learn a set of parameters at the input level, and Adapters~\cite{uniadapter,adapter}
learn to adjust intermediate representations, both for better adapting
large-scale models for specific tasks. Although deployment prior resembles
a kind of understandable prompt, our detector predicts a part of the
model parameters according to it.

\textbf{Graph knowledge} including scene graph~\cite{scenegraph},
knowledge graph~\cite{object_kg}, and even label graph~\cite{labelrelation}
can aid object detection. Most of these graphs are used in a detector
with hand-craft rules~\cite{object_kg} or highly entangled modeling~\cite{labelrelation},
and require external data. Conversely, our graph is merely a representation
of the conditional probability statistics among classes, which does
not rely on any external data than the detection dataset.

Despite slight resemblance to other topics including \textit{domain adaptation}~\cite{DA_prompt,DA_univeral_objectdetection,object_relation_cross_domin}
and \textit{test-time adaptation}~\cite{test_time_adaptation,tta2,cotta,vsoda,Kim_tta},
the proposed method is different. We focus on explicitly modeling
the domain shift, which is the object relations in this paper, and
build a framework to calibrate the test-time behavior of the detector
without any back-propagation.

\section{Calibratable Object Detector}

\label{sec:3}%
The proposed method is built upon DINO~\cite{dino}, which casts
object detection as set prediction~\cite{detr,deformabledetr} using
Transformer~\cite{transformer}. Given an object detection dataset
with $K$ object classes, comprising image-label pairs: $(x_{\text{img}},y)$,
where $x_{\text{img}}\in\mathbb{R}^{3\times H_{\text{img}}\times W_{\text{img}}}$
is an RGB image, and $y$ is a set containing $N$ pairs of class
and bounding box annotations $y=\{(c_{i},b_{i})|i=1,\ldots,N\}$.
Denote $\phi_{x}(x_{\text{img}})$ as the backbone, which computes
the feature map $x$ of shape $(C,H,W)$. The Transformer takes $x$
as input sequence and yields decoder outputs $\{h_{m}\}_{m=1}^{M}$
of length $d$ for each of the $M$ object queries. The $h_{m}$ is
used to predict scores and bounding box.

{} Beside the image, consider a graph $\langle V,E\rangle$ as an additional
input. Specifically, the nodes $V\in\mathbb{R}^{d\times K}$ are per-class
embeddings. The edge $E\in[0,1]^{K\times K}$ is a conditional probability
matrix for every pair of classes (object relations). The $(i,j)$-th
element in $E$ denotes the possibility that object class $i\in\{1,\ldots,K\}$
appears given the presence of object class $j\in\{1,\ldots,K\}$,
\ie, $P(i|j)$. The choice of conditional probability is based on
the observed consistency between it and input pixel gradient attribution.
The graph explicitly describes the relationship among classes.

Then, a calibratable object detector $f(x_{\text{img}};V,E)$ predicts
a set of label-box pairs with its behavior bound to $V$ and $E$.
Given the deployment prior $E\in[0,1]^{K\times K}$, its injection
is achieved as edge editing at run-time, which helps the detector
better adapt to object relation changes. The closer the injected prior
is to the real shifted relations, the better the detector performance
is expected to be. The overview of the proposed framework can be found
in Fig.~\ref{fig:architecture}. We abbreviate the proposed framework
as ``CaliDet''.

As the calibration of $V$ is conceptually incremental or transfer
learning, which is beyond the scope of this paper, we specifically
focus on the calibration of $E$ (object relations). Therefore, we
simplify the CaliDet notation as $f(x_{\text{img}};E)$.

\subsection{Calibratable Edges $E$ for Node Interaction}

\label{sec:31}

The key idea for prior injection is to predict the class center shift
based on $\langle V,E\rangle$. If we directly model the interactions
among the nodes $V$ with \eg, Transformer, the model implicitly
learns fixed relations, which is not calibratable at run-time. Instead,
we model the node interactions using a modified encoder-only Transformer
named ``CaliFormer''. 

\textbf{CaliFormer} leverages $E$ as a self-attention bias to interfere
with the interactions among nodes $V$, and predicts the calibration
vectors $V'$ for class center shifts. In particular, we first convert
the deployment prior into a difference matrix $\Delta E\triangleq E-E_{0}\in[-0.5,0.5]^{K\times K}$.
The $E_{0}$ denotes the constant flat prior, for the case where there
is no prior at all.

Subsequently, we modify the Transformer Encoder~\cite{transformer},
where $\Delta E$ is used as a bias for the self-attention among $V$:
\begin{equation}
\boldsymbol{A}=\text{softmax}(\frac{\boldsymbol{Q}\boldsymbol{K}^{\mathsf{T}}}{\sqrt{d_{k}}}+\Delta E^{\mathsf{T}})\boldsymbol{V},\label{eq:califormer}
\end{equation}
where $\boldsymbol{Q}$, $\boldsymbol{K}$, $\boldsymbol{V}$ are
the query, key, and value, respectively, as defined in~\cite{transformer}.
This is a common~\cite{graphomer,plain-detr} modification. 

Since $\Delta E$ interferes with the self-attention among $V$, it
can be seen as a hint for increasing or decreasing the inclusion of
cues of another class into the corresponding embedding. For example,
denoting $\Delta P(i|j)$ as the $(i,j)$-th element of $\Delta E$,
a positive $\Delta P(i|j)$ informs the model to include more cues
of class $i$ into class $j$, or reduce the dependency of class $j$
to class $i$ by a negative $\Delta P(i|j)$. The case when $\Delta P(i|j)=0$
means no deployment prior is provided. In this way, CaliFormer $g(\cdot)$
transduces the sequence of the columns of $V$ into the per-class
calibration vectors $V'\triangleq g(V,\Delta E)$. No position encoding
is needed here, because the nodes $V$ are permutation invariant.

After obtaining the calibration vectors $V'$, we incorporate them
into the prediction heads of the object detector as the class center
shifts, in order to adjust the detector behavior.

\textbf{Prediction Head} leverges a single linear layer for classification~\cite{deformabledetr,dino,detr}.
Let the linear layer be $\phi_{c}(h_{m})=W^{\mathsf{T}}\cdot h_{m}+b$,
where $W\in\mathbb{R}^{d\times K}$ is the weight, and $b\in\mathbb{R}^{d}$
is the bias. As the weights can be seen as object class centers~\cite{arcface},
we add the calibration vectors $V'$, scaled by a constant factor
$\rho$, to the weights to reflect the class center shifts:
\begin{equation}
\phi_{c}(h_{m})=(W+\rho V')^{\mathsf{T}}\cdot h_{m}+b,\label{eq:prediction-head}
\end{equation}
for dynamically adjusting the class centers based on the injected
prior $\Delta E$. The constant $\rho$ is a scaling factor. The prediction
head for bounding boxes is left unchanged.

\subsection{Binding Detector Behavior to $E$}

\label{sec:32}

Assume the data distribution $\mathcal{X}(E)$ is parametrized by
the edge $E$. Let $E_{0}$ be the flat prior\footnote{In $E_{0}$, all off-diagonal values are $0.5$. The diagonal values
are $1$.}, $E_{t}$ be the statistics of training set, and $E_{v}$ for validation
set. $E_{x}$ is the conditional probability matrix estimated\footnote{For each object class $j$ present in the image, $P(i|j)$ is set
to $1$ when class $i$ is present, or $0$ when class $i$ is absent.
For each class $j$ absent from the image, $P(i|j)$ will be inherited
from flat prior $E_{0}$ as nothing is known.} using a single data point $x$. Similarly, $E_{b}$ is estimated
using a batch of data points. For example, $E_{b(1)}=E_{x}$ where
the batch size is one.

Let $\mathscr{L}(\cdot)$ be the sum of classification loss and bounding
box loss for a single training sample, a typical detector $f_{\theta^{*}}(\cdot)$
is trained to minimize the empirical risk on the training set:
\begin{equation}
\theta^{*}=\arg\min_{\theta}\mathbb{E}_{x\sim\mathcal{X}(E_{t})}\{\mathscr{L}[f_{\theta}(x)]\}.\label{eq:stdopt}
\end{equation}
In case of a distribution shift from $\mathcal{X}(E_{t})$ to $\mathcal{X}(E_{v})$,
the model might perform slightly worse, as the model has implicitly
learned the training set statistics $E_{t}$:
\begin{equation}
\mathbb{E}_{x\sim\mathcal{X}(E_{v})}\{\mathscr{L}[f_{\theta^{*}}(x)]\}\geqslant\mathbb{E}_{x\sim\mathcal{X}(E_{t})}\{\mathscr{L}[f_{\theta^{*}}(x)]\}.\label{eq:dshift}
\end{equation}
Given the deployment prior, we use a calibratable detector $f_{\theta^{\dagger}}(x;E)$.
The more accurate the given edge $E$ is for describing the data,
the more accurate the detector is expected to be. This can be expressed
as the following inequality:
\begin{align}
 & \mathbb{E}_{x\sim\mathcal{X}(E_{t})}\big\{\mathscr{L}[f_{\theta^{\dagger}}(x;E_{0})]\big\}\nonumber \\
\geqslant\; & \mathbb{E}_{x\sim\mathcal{X}(E_{t})}\big\{\mathscr{L}[f_{\theta^{\dagger}}(x;E_{t})]\big\}\label{eq:ineq}\\
\geqslant\; & \mathbb{E}_{x\sim\mathcal{X}(E_{t})}\big\{\mathscr{L}[f_{\theta^{\dagger}}(x;E_{x})]\big\}.\nonumber 
\end{align}
When a model can satisfy this inequality, it is expected to perform
better on $\mathcal{X}(E_{v})$ with $f_{\theta^{\dagger}}(x;E_{v})$.
Assuming that the model $f_{\theta^{*}}(\cdot)$ implicitly learns
the $E_{t}$ from $\mathcal{X}(E_{t})$, then{\small
\begin{align}
\mathbb{E}_{x\sim\mathcal{X}(E_{v})}\big\{\mathscr{L}[f_{\theta^{*}}(x)]\big\}\approx & \mathbb{E}_{x\sim\mathcal{X}(E_{v})}\big\{\mathscr{L}[f_{\theta^{\dagger}}(x;E_{t})]\big\}\nonumber \\
\geqslant & \mathbb{E}_{x\sim\mathcal{X}(E_{v})}\big\{\mathscr{L}[f_{\theta^{\dagger}}(x;E_{v})]\big\}.\label{eq:whatif}
\end{align}
}

These require the detector behavior to be bound to $E$. While the
deployment prior is injected into the model during training through
the proposed architecture, the standard training loss is invariant
to $E$, unable to guide the model to correctly utilize the information
in $E$. So, we need to change the loss landscape with respect to
$E$, by introducing an extra loss term to pose a higher penalty when
the given $E$ is more accurate, in order to implicitly enforce Eq.~\eqref{eq:ineq}. 

\begin{figure}
\includegraphics[width=1.0\columnwidth]{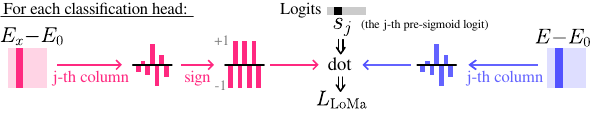}

\vspace{-0.5em}

\caption{Demonstration of Logit Manipulation Loss (LoMa).}

\label{fig:epr}

\vspace{-0.5em}
\end{figure}

\textbf{Logit Manipulation (LoMa).} Denote $s_{j}$ as pre-sigmoid
logit for the class $j$ at the classification head corresponding
to decoder output $h_{m}$. Recall that the $j$-th column in $\Delta E=E-E_{0}$
denotes whether class $j$ should depend more or less on each class
$i$. Therefore, we use the mean value of the $j$-th column in $\text{sign}(E_{x}-E_{0})\odot(E-E_{0})$
to measure whether the arbitrary injected $E$ is accurate or not,
where ``$\odot$'' is element-wise product. Thus, the negative of
this mean value is multiplied to $s_{j}$ as the penalty for class
$j$, encouraging $s_{j}$ to be higher when $E$ is accurate, or
lower when inaccurate:
\begin{equation}
L_{j}=s_{j}\cdot\frac{1}{K}\sum_{i}[-\text{sign}(E_{x}-E_{0})^{(i,j)}\cdot(E-E_{0})^{(i,j)}].
\end{equation}
 The term $L_{j}$ is averaged over all classes as the LoMa loss for
a single classification head: $L_{\text{LoMa}}=\gamma\sum_{j}L_{j}/K$,
as shown in Fig.~\ref{fig:epr}. The constant $\gamma$ is a balancing
hyper-parameter .

In this way, we reshape the loss landscape to peak at $E_{x}$ where
we expect better performance according to Eq.~\eqref{eq:ineq}. And
the loss is lower if we inject a misleading $E$.

\textbf{Edge Sampling.} A proper choice of $E$ for training is vital
for generalization against different $E$ at run-time. To formulate
the training objective, we first rewrite the original training objective
in an equivalent double-expectation form:
\begin{equation}
\theta^{*}=\arg\min_{\theta}\mathbb{E}_{E\sim\delta(E_{t})}\big\{\mathbb{E}_{x\sim\mathcal{X}(E)}\{\mathscr{L}[f_{\theta}(x)]\}\big\},\label{eq:stdoptext}
\end{equation}
with $\delta(E_{t})$ denoting the Dirac function. It shows the training
of $\theta^{*}$ is invariant to $E$. Replacing $\delta(E_{t})$
with an edge sampling distribution $\mathcal{E}$ will make the parameters
optimized against varying $E$. Namely, the training of CaliDet aims
to simultaneously generalize against different $E$ and $x$:
\begin{align}
\theta^{\dagger} & =\arg\min_{\theta}\mathbb{E}_{E\sim\mathcal{E}}\big\{\mathbb{E}_{x\sim\mathcal{X}(E_{t})}\{\mathscr{L}[f_{\theta}(x;E)]\}\big\}.\label{eq:caliopt}
\end{align}
The dataset term is fixed at $\mathcal{X}(E_{t})$ instead of $\mathcal{X}(E)$
because we only use one training dataset in practice.

Inspired by meta learning~\cite{maml}, we regard different $E$
as different ``meta tasks''. We empirically define the distribution
$\mathcal{E}$ by the following algorithm: (1) randomly select a prior
from $\{E_{x},E_{b},E_{t}\}$ with equal probability; (2) generate
a noise matrix from $\mathcal{N}(0,\sigma^{2})$, and add it to the
selected prior; (3) clip the prior to $[0,1]$, and reset its diagonal
to $1$.

The $E$ sampled from $\mathcal{E}$ is used as an input to $f_{\theta^{\dagger}}(\cdot;\cdot)$
for training, and re-calculated for every mini-batch. The Gaussian
noise facilitates generalization by covering more different $E$,
and prevent overfitting at $E_{x}$, $E_{b}$ and $E_{t}$.

\subsection{Inference and Run-time Calibration}

\label{sec:34}

During the inference stage, deployment prior is injected to the model
via the input $E$ as $f_{\theta^{\dagger}}(x;E)$. When there is
no prior, we use $E=E_{t}$ (training set statistics) as the default.
Given an arbitrary set of $x$, the result of $f_{\theta^{\dagger}}(x;E)$
varies according to the given $E$. The more accurately that $E$
describes the distribution from which $x$ is drawn, the more accurate
the result should be. Since the injection process is merely changing
the $K\times K$ input matrix $E$, while leaving model parameters
intact, it is called ``run-time'' calibration.

\textbf{Self-Calibration.} In practice, the deployment prior is not
always available. In the worst case, the object relations in the new
distribution are completely unknown. This does not mean the detector
is limited by the assumption that the deployment prior is available.
In contrast, the detector can still automatically find an approximate
prior to calibrate itself. This algorithm is summarized in Algo.~\ref{algo:self-calibrate}.

Given an arbitrary set of sample images $X$, we first initialize
the self-calibrated edge $E_{c}$ as the default prior $E_{t}$, because
the object relation shift with respect to $E_{t}$ in the underlying
distribution of $X$ is unknown.

Next, we forward the model $f_{\theta^{\dagger}}(x;E_{c})$ on the
given set $X$, and calculate the conditional probability matrix $E_{i}$
purely based on the model predictions. Before we move $E_{c}$ towards
$E_{i}$ with a step size $\eta$, it must be pointed out that an
imperfect detector $f_{\theta^{\dagger}}(x;E_{c})$ will expectedly
produce false predictions, which could mislead the detector.

Hence, a weight matrix $Z\in\mathbb{R}^{K\times K}$ is introduced.
We first calculate a vector $z\in\mathbb{R}^{1\times K}$ corresponding
to every $x\in X$, where the $i$-th element of $z$ is the maximum
confidence score among the predictions for class $i$, or $0$ when
class $i$ is not predicted from $x$. Then, we figure out the average
$z$, and repeat it to satisfy the shape of $Z$. As a result, the
$E_{c}$ update is mainly guided by confident predictions, lest the
detector be quickly misled by its own faulty predictions. We iteratively
update $E_{c}$ with a frozen CaliDet when no prior is available.
The clip operation ensures that $E_{c}$ remains a matrix with valid
conditional probability values.

In this way, a frozen model can still obtain a performance gain upon
an arbitrary test distribution shift for ``free'' (\ie, without
any parameter update or data annotation).

\begin{algorithm}[t]
\textbf{~~~Input:} Arbitrary set of samples $X$ and step size $\eta$. \\
\textbf{Output:} Self-calibrated edge $E_c \in [0,1]^{K\times K}$.
\begin{algorithmic}[1]
\State $E_c \gets E_t$ \Comment{Initialize $E_c$ as the default prior}
\For{$i \gets 1,2,\ldots,$ MaxIteration}
  \State $E_i\gets$ Statistics\big\{$f_{\theta^{\dagger}}(x;E_c)$ for all $x\in X$ \big\}
  \State $Z \gets$ Repeat(Mean\{$z$ for all $x \in X$\})
  \State $E_c \gets$ Clip\big\{ $E_c + \eta Z \cdot (E_i - E_c)$, $[0,1]^{K\times K}$\big\}
\EndFor
\end{algorithmic}

\caption{Self-Calibration with $E_{c}$}

\label{algo:self-calibrate}
\end{algorithm}

\section{Experiments}

\label{sec:40}

In this paper, we aim to endow a detector with the run-time calibration
capability without sacrificing the default AP (no injection) compared
to the baseline detector. The experiments are conducted with eight
NVIDIA RTX A5000 GPUs. Our code is based on PyTorch~\cite{pytorch}
and DINO~\cite{dino}.

\textbf{Dataset.} To validate our method, we conduct experiments on
the COCO 2017 dataset~\cite{coco}, following \cite{deformabledetr,detr,dino}.
It contains 118K training images and 5K validation images.

\textbf{Baseline.} We adopt DINO~\cite{dino} (4 scales, ResNet-50~\cite{resnet}
backbone, 12 epochs) as the not calibratable baseline. Unless explicitly
mentioned, we retain all overlapping model details. Note, the proposed
method is not specific to DINO. See supplementary for experiments
on D-DETR~\cite{deformabledetr}.

\textbf{Training.} To accelerate the training process, we initialize
our model from the officially pre-trained DINO~\cite{dino}. The
learning rate for the pre-trained parts is decayed by $0.1$ at the
beginning. The training scheme is $6$ epochs with learning rate decay
after $5$ epochs, in order to roughly align our model's default AP
(\ie, $E=E_{t}$) with the baseline AP.

\textbf{Tunables.} The embedding dimension $d=256$ for $V$ is also
the Transformer dimension. The nodes $V$ are initialized with $\mathcal{N}(0,0.01^{2})$
following~\cite{arcface}. The CaliFormer consists of $3$ layers
of Transformer Encoder with the attention bias described in Sec.~\ref{sec:31}.
We set the scaling factor of $V'$ as $\rho=0.2$; the balancing parameter
$\gamma=20$ for $L_{\text{LoMa}}$; the Gaussian variance as $\sigma^{2}=0.16^{2}$
for $\mathcal{E}$; the step size $\eta=4.0$ for self-calibration.
We use a mini-batch of size 2 following~\cite{dino}.

\subsection{Evaluation Protocol for Calibratable Detector}

\label{sec:41}

\textbf{Standard Evaluation (Sec.~\ref{sec:42}).} To verify if the
proposed method can adapt to distribution shifts at run-time, it is
evaluated against different priors, ranging from inaccurate ones to
accurate ones: $\overline{E_{x}}$, $E_{0}$, $E_{t}$, $E_{v}$,
$E_{b}$, $E_{x}$. The $\overline{E_{x}}$ is the result of flipping
$0$ and $1$ in $E_{x}$ except for the diagonal, and hence is the
most misleading one. For the validation dataset $\mathcal{X}(E_{v})$,
we expect the Average Precision (AP)~\cite{detr,deformabledetr}
to be higher if the provided $E$ is accurate, or lower if misleading.
The $E_{t}$ is regarded as the default deployment prior. And the
AP when injecting $E_{x}$ can provide an empirical performance upper
bound with prior injection according to Eq.~(\ref{eq:ineq}). If
the model performance follows the expected order, it means the model
has generalized for different edges covered by $\mathcal{E}$.

\textbf{Subset Evaluation (Sec.~\ref{sec:43}).} In well-controlled
datasets, the test distribution may not significantly deviate from
$E_{t}$. As $E_{v}\approx E_{t}$, the performance gain by injecting
$E_{v}$ may be negligible. To simulate some shifts from $E_{v}$,
we randomly segment the validation set into a series of equal-sized
non-overlapping subsets, and inject the corresponding subset statistics
$E_{b}$. Note, in this case, $E_{b}$ is still a random matrix centered
at $E_{v}$, but along with a notable variance. The performance averaged
over subsets is expected to be higher with $E_{b}$ compared to $E_{t}$.
Note, according to the Law of Large Numbers, the conditional probability
statistics calculated from a larger subset will be more approximate
to $E_{v}$. Thus, very large subsets are not necessary for subset
evaluation. 

\textbf{Self-Calibration (Sec.~\ref{sec:44}).} In the subset evaluation,
we use the subset statistics $E_{b}$ as the deployment prior. However,
those statistics are not always available in practice as the statistics
rely on annotations. Thus, we use the model's own predictions to calibrate
itself on random subsets, and show the performance curves during the
process of Algo.~\ref{algo:self-calibrate}.

\subsection{Standard Evaluation w/ Different Priors}

\label{sec:42}

As a prerequisite for a valid calibratable detector, its detection
performance by the standard COCO metrics should be at least on par
with the baseline. Thus, we compare the baseline model to our model
with the default deployment prior in these metrics in Tab.~\ref{tab:dino-const}.
Meanwhile, we inject different priors to verify whether the model
can leverage the injected prior. Note, the injection process only
changes the input edge $E$.

The baseline achieves $49.07$ overall AP, while our model achieves
an on-par performance $49.27$ with the default prior $E_{t}$ inherited
from the training set. When we provide the flat prior $E_{0}$, which
is less accurate than $E_{t}$, and represents the case when there
is no prior knowledge at all, the model performs slightly worse by
$1.30$ AP. If we provide an even worse prior $\overline{E_{x}}$
to intentionally ``mislead'' the model, the performance further
drops to $35.83$.

\begin{table}
\resizebox{1.0\linewidth}{!}{%
\setlength{\tabcolsep}{4pt}%
\begin{tabular}{cccccccc}
\toprule 
\multirow{2}{*}{\textbf{Method}} & \multirow{2}{*}{\textbf{Injection}} & \multicolumn{6}{c}{\textbf{Standard COCO Metrics}}\tabularnewline
\cmidrule{3-8}
 &  & AP & AP$_{50}$ & AP$_{75}$ & AP$_{S}$ & AP$_{M}$ & AP$_{L}$\tabularnewline
\midrule
\midrule 
DINO~\cite{dino} & - & 49.07 & 66.72 & 53.57 & 32.67 & 52.36 & 63.04\tabularnewline
\midrule 
\rowcolor{blue!7}\cellcolor{white} & $\overline{E_{x}}$ & 35.83 & 46.48 & 39.10 & 18.14 & 38.07 & 53.24\tabularnewline
 & $E_{0}$ & 47.97 & 64.50 & 52.37 & 30.79 & 51.65 & 62.49\tabularnewline
\rowcolor{black!7}\cellcolor{white}CaliDet & $E_{t}$ & 49.27 & 66.73 & 53.80 & 32.75 & 52.57 & 63.50\tabularnewline
(DINO) & $E_{v}$ & 49.30 & 66.77 & 53.84 & 32.77 & 52.58 & 63.53\tabularnewline
 & $E_{b}$ & 51.69 & 70.62 & 56.48 & 35.82 & 55.08 & 65.29\tabularnewline
\rowcolor{magenta!7}\cellcolor{white} & $E_{x}$ & 51.94 & 70.92 & 56.69 & 36.10 & 55.27 & 65.76\tabularnewline
\bottomrule
\end{tabular}

}
\vspace{0.1em}

\caption{Standard COCO Evaluation Results. The AP given the default deployment
prior $E_{t}$ is close to that of the baseline method. Compared to
$E_{t}$, the less accurate edges $\overline{E_{x}}$ and $E_{0}$
lead to lower performance, while more accurate edges such as $E_{x}$
lead to higher performance. Note, as shown in Fig.~\ref{fig:pcolor-e},
the difference between $E_{v}$ and $E_{t}$ is marginal to make a
larger difference in AP. Likewise, the difference between $E_{b}$
and $E_{x}$ is also small because the mini-batch size is $2$ for
DINO. Namely $E_{b(2)}\approx E_{x}$.}

\label{tab:dino-const}
\end{table}

\begin{figure}
\includegraphics[width=\columnwidth]{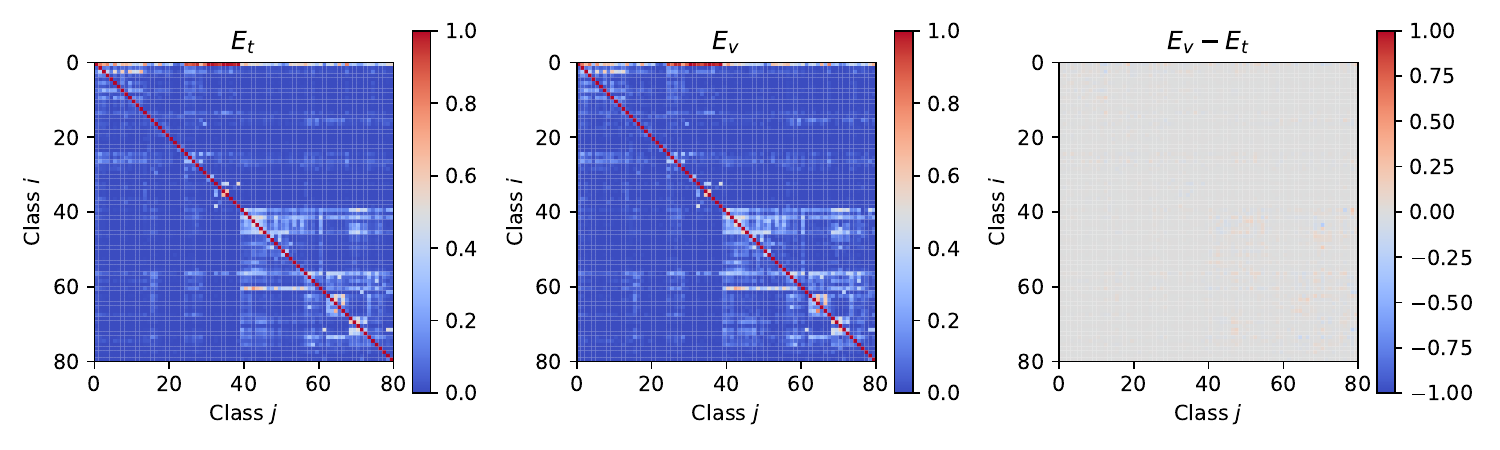}

\caption{Visualization of $E_{t}$, $E_{v}$, and $E_{v}-E_{t}$. The mean
absolute error between the two matrices is $\epsilon=\sum\text{abs}(E_{v}-E_{t})/K^{2}=0.008$.
The 0-th, 50-th, 90-th, 97-th, 100-th percentile values of $\text{abs}(E_{v}-E_{t})$
are $0.0,$ $0.003,$ $0.021$, $0.044$, $0.272$, respectively.}

\label{fig:pcolor-e}
\end{figure}

If we provide the $E_{v}$ as the prior, the performance slightly
improves compared to the $E_{t}$ case. This is because, on the well-controlled
COCO dataset, the difference between $E_{t}$ and $E_{v}$ is too
marginal to make a larger difference in AP, as shown in Fig.~\ref{fig:pcolor-e}.
Notably, the difference between $E_{v}$ and $E_{t}$ is even smaller
than the Gaussian noise used for the edge sampling during the training
process. This observation is exactly our motivation for the subset
evaluations with a much larger simulated relation shift.

Conversely, if we provide a more accurate prior, such as the mini-batch
prior $E_{b}$ and the per-sample prior $E_{x}$ (most accurate),
the performance will be improved to $51.69$ and $51.94$. Note, the
difference between $E_{b}$ and $E_{x}$ is small because the mini-batch
size is $2$ for DINO. Namely $E_{b(2)}\approx E_{x}$ in this case.
Although it may be impossible to obtain $E_{x}$ upon deployment,
the corresponding performance can be regarded as the empirical upper
bound of the performance gain through calibration. The experiments
demonstrate that the detector behavior is effectively bound to $E$.

In summary, the experiments in Tab.~\ref{tab:dino-const} demonstrate
that: (1) our method does not sacrifice performance compared to the
baseline method, unless we intentionally mislead the model with a
wrong prior; (2) All the AP metrics follow the expected order, that
a more accurate edge $E$ leads to a higher AP, while a less accurate
edge leads to a lower AP. This reflects our expectation in Eq.~(\ref{eq:ineq})
and Eq.~(\ref{eq:whatif}). Empirically, this is a sanity check when
tuning the hyper-parameters when adopting CaliDet to a different baseline
detector.

\begin{table}
\resizebox{1.0\linewidth}{!}{%
\setlength{\tabcolsep}{2pt}%

\begin{tabular}{ccccccccc}
\toprule 
\multirow{2}{*}{\textbf{Subset Size}} & \multirow{2}{*}{\textbf{Injection}} & \multirow{2}{*}{$\epsilon$} & \multicolumn{6}{c}{\textbf{Metrics Averaged over COCO Subsets}}\tabularnewline
\cmidrule{4-9}
 &  &  & \multicolumn{1}{c}{\cellcolor{red!7}AP} & \multicolumn{1}{c}{AP$_{50}$} & \multicolumn{1}{c}{\cellcolor{red!7}AP$_{75}$} & \multicolumn{1}{c}{AP$_{S}$} & \multicolumn{1}{c}{\cellcolor{red!7}AP$_{M}$} & \multicolumn{1}{c}{AP$_{L}$}\tabularnewline
\midrule
\midrule 
\multirow{3}{*}{8} & \cellcolor{black!7}$E_{t}$ & 0 & 62.93 & 79.69 & 67.56 & 45.90 & 66.41 & 79.77\tabularnewline
 & $E_{b(8)}$ & 0.331 & 63.15 & 80.01 & 67.76 & 46.10 & 66.66 & 79.87\tabularnewline
 &  &  & {\footnotesize\textcolor{magenta}{(+ 0.22)}} & {\footnotesize\textcolor{magenta}{(+ 0.32)}} & {\footnotesize\textcolor{magenta}{(+ 0.20)}} & {\footnotesize\textcolor{magenta}{(+ 0.20)}} & {\footnotesize\textcolor{magenta}{(+ 0.25)}} & {\footnotesize\textcolor{magenta}{(+ 0.10)}}\tabularnewline
\midrule
\multirow{3}{*}{16} & \cellcolor{black!7}$E_{t}$ & 0 & 61.06 & 77.63 & 65.52 & 44.40 & 64.65 & 78.21\tabularnewline
 & $E_{b(16)}$ & 0.276 & 61.31 & 77.96 & 65.77 & 44.56 & 64.83 & 78.44\tabularnewline
 &  &  & {\footnotesize\textcolor{magenta}{(+ 0.25)}} & {\footnotesize\textcolor{magenta}{(+ 0.33)}} & {\footnotesize\textcolor{magenta}{(+ 0.25)}} & {\footnotesize\textcolor{magenta}{(+ 0.16)}} & {\footnotesize\textcolor{magenta}{(+ 0.18)}} & {\footnotesize\textcolor{magenta}{(+ 0.23)}}\tabularnewline
\midrule
\multirow{3}{*}{32} & \cellcolor{black!7}$E_{t}$ & 0 & 59.74 & 76.05 & 64.17 & 42.70 & 63.16 & 76.85\tabularnewline
 & $E_{b(32)}$ & 0.202 & 60.00 & 76.39 & 64.44 & 42.82 & 63.43 & 77.00\tabularnewline
 &  &  & {\footnotesize\textcolor{magenta}{(+ 0.26)}} & {\footnotesize\textcolor{magenta}{(+ 0.34)}} & {\footnotesize\textcolor{magenta}{(+ 0.17)}} & {\footnotesize\textcolor{magenta}{(+ 0.12)}} & {\footnotesize\textcolor{magenta}{(+ 0.27)}} & {\footnotesize\textcolor{magenta}{(+ 0.15)}}\tabularnewline
\midrule
\multirow{3}{*}{64} & \cellcolor{black!7}$E_{t}$ & 0 & 57.89 & 74.21 & 62.41 & 40.94 & 61.69 & 75.07\tabularnewline
 & $E_{b(64)}$ & 0.122 & 58.13 & 74.54 & 62.67 & 41.07 & 61.96 & 75.20\tabularnewline
 &  &  & {\footnotesize\textcolor{magenta}{(+ 0.24)}} & {\footnotesize\textcolor{magenta}{(+ 0.33)}} & {\footnotesize\textcolor{magenta}{(+ 0.26)}} & {\footnotesize\textcolor{magenta}{(+ 0.13)}} & {\footnotesize\textcolor{magenta}{(+ 0.27)}} & {\footnotesize\textcolor{magenta}{(+ 0.13)}}\tabularnewline
\midrule
\multirow{3}{*}{128} & \cellcolor{black!7}$E_{t}$ & 0 & 55.74 & 72.16 & 60.32 & 39.17 & 60.04 & 72.89\tabularnewline
 & $E_{b(128)}$ & 0.062 & 55.89 & 72.35 & 60.47 & 39.25 & 60.12 & 72.98\tabularnewline
 &  &  & {\footnotesize\textcolor{magenta}{(+ 0.15)}} & {\footnotesize\textcolor{magenta}{(+ 0.19)}} & {\footnotesize\textcolor{magenta}{(+ 0.15)}} & {\footnotesize\textcolor{magenta}{(+ 0.08)}} & {\footnotesize\textcolor{magenta}{(+ 0.08)}} & {\footnotesize\textcolor{magenta}{(+ 0.09)}}\tabularnewline
\midrule
\multirow{3}{*}{256} & \cellcolor{black!7}$E_{t}$ & 0 & 53.67 & 70.33 & 58.26 & 37.62 & 57.77 & 71.09\tabularnewline
 & $E_{b(256)}$ & 0.034 & 53.82 & 70.50 & 58.44 & 37.70 & 57.94 & 71.14\tabularnewline
 &  &  & {\footnotesize\textcolor{magenta}{(+ 0.15)}} & {\footnotesize\textcolor{magenta}{(+ 0.17)}} & {\footnotesize\textcolor{magenta}{(+ 0.18)}} & {\footnotesize\textcolor{magenta}{(+ 0.08)}} & {\footnotesize\textcolor{magenta}{(+ 0.17)}} & {\footnotesize\textcolor{magenta}{(+ 0.05)}}\tabularnewline
\bottomrule
\end{tabular}

}
\vspace{0.1em}

\caption{Subset Evaluation with Varying Subset Size. We split the validation
datasets into subsets with varying sizes, and report the metrics averaged
over all subsets of the respective sizes. CaliDet shows a ``free''
(\ie, without any parameter update) performance gain as long as a
more accurate prior than $E_{t}$ is provided.}

\label{tab:dino-mpi}
\end{table}

\subsection{Subset Evaluation (Simulated Shifts)}

\label{sec:43}

In the previous subsection, we note the difference between $E_{v}$
and $E_{t}$ is marginal on COCO, a well-controlled dataset. This
does not lead to a clear difference when $E_{v}$ is injected compared
to to default prior, \ie, $E_{t}$. Thus, we split the validation
set into subsets with varying sizes to simulate the distribution shift,
and report the metrics averaged over all subsets, as shown in Tab.~\ref{tab:dino-mpi}.
Note, the injection only changes the input edge $E$, with model parameters
being frozen.

With a small subset size (\eg, 8), the corresponding subset statistics
$E_{b(8)}$ will clearly deviate from $E_{t}$, as suggested by its
mean absolute error $\epsilon$ with respect to $E_{t}$. This can
lead to a clear performance gain when we inject the corresponding
priors. For instance, the AP increases from $62.93$ ($E_{t}$) to
$63.15$ with the subset size $8$ and the subset statistics $E_{b(8)}$
as the injected prior. The results for the other subset sizes also
suggest a performance gain compared to $E_{t}$, as long as the corresponding
$E_{b}$ clearly differs from $E_{t}$.

As discussed in Sec.~\ref{sec:41}, a larger subset size will make
the subset statistics $E_{b}$ closer to $E_{v}$, while $E_{v}\approx E_{t}$.
When the subset size is 256, the injection of $E_{b(256)}$ leads
to a less notable difference in performance, because it is very close
to $E_{v}$ as $\epsilon=0.034$. We refrain from experimenting on
larger subsets because they do not simulate clear distribution shifts.

In summary, the subset evaluation for simulating distribution shifts
by relation change demonstrates the effectiveness of the proposed
method. As long as the prior is more accurate than $E_{t}$, a consistent
performance gain is expected for ``free'', \ie, with merely changes
in the input matrix $E$ while keeping the whole model frozen. Thus,
CaliDet is effective.

\subsection{Self-Calibration Evaluation}

\label{sec:44}

In practice, the deployment prior is not always available. Thanks
to the statistically meaningful definition of object relations, the
CaliDet can be designed to calibrate itself using its own prediction
results following Algo.~\ref{algo:self-calibrate}. Specifically,
we sample subsets of varying sizes from COCO validation set. The subsets
of each size sum up to 1024 samples. During self-calibration, each
individual subset is regarded as the set of images samples $X$ used
Algo.~\ref{algo:self-calibrate}, and the model is frozen.

\begin{figure}
\includegraphics[width=\columnwidth]{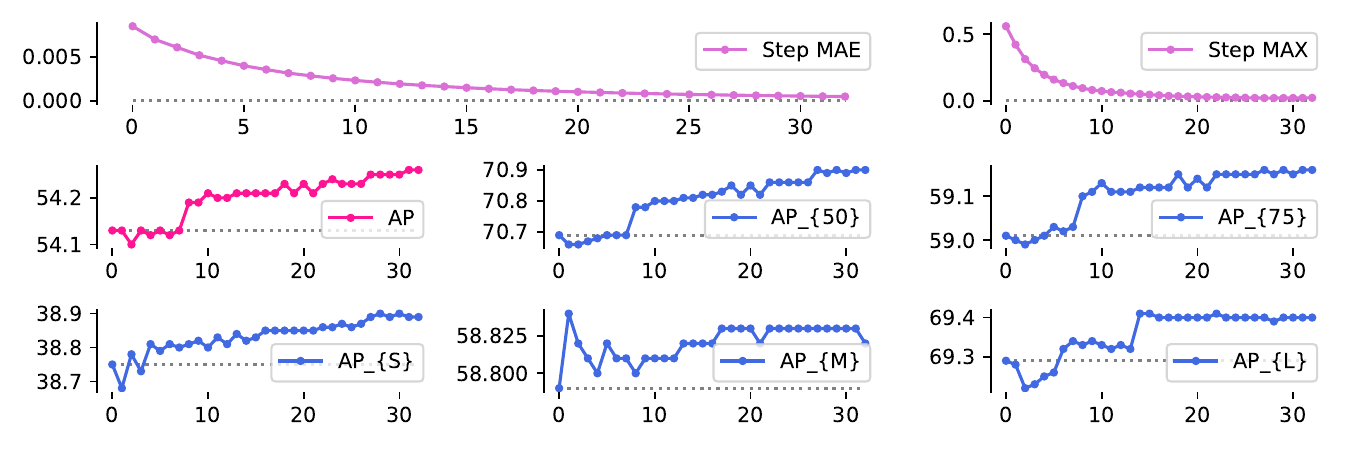}

\vspace{-0.7em}

\caption{Self Calibration Results with Subset Size 256.}

\label{fig:sc256}
\end{figure}

\begin{figure}
\includegraphics[width=\columnwidth]{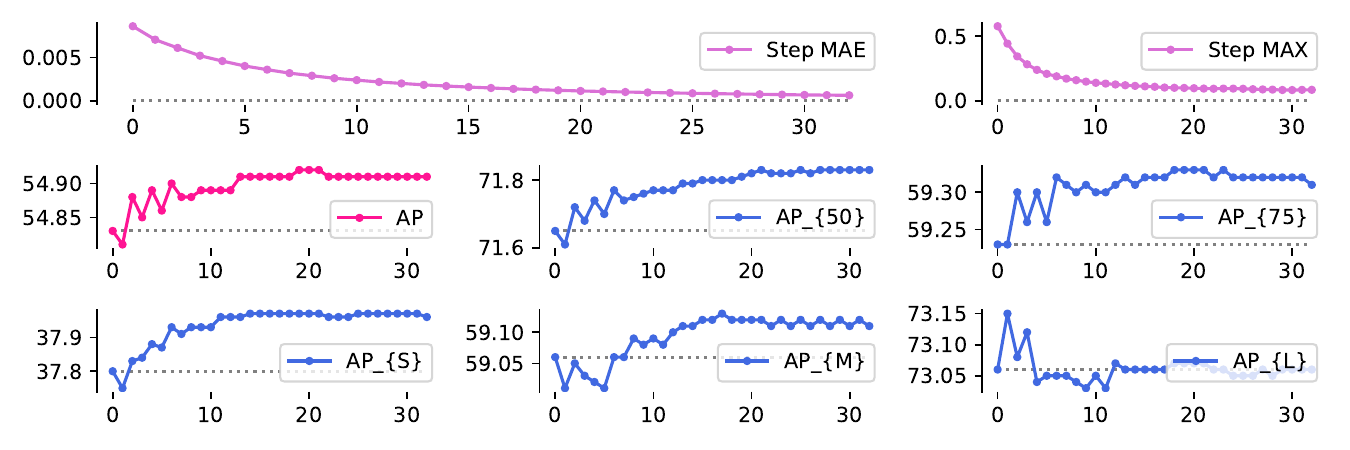}

\vspace{-0.7em}

\caption{Self Calibration Results with Subset Size 128.}

\label{fig:sc128}
\end{figure}

\begin{figure}
\includegraphics[width=\columnwidth]{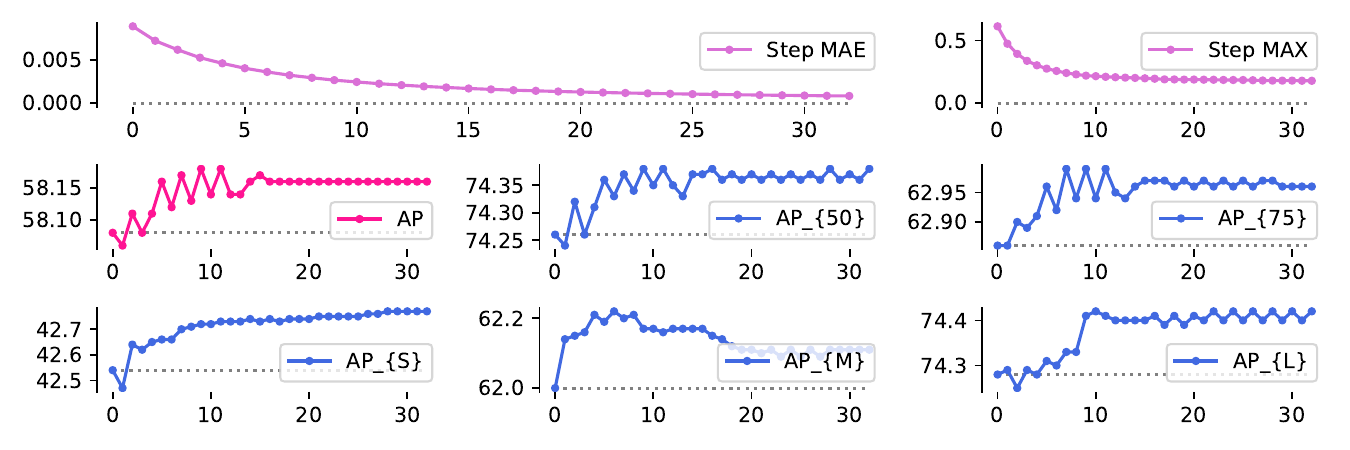}

\vspace{-0.7em}

\caption{Self Calibration Results with Subset Size 64.}

\label{fig:sc64}
\end{figure}

\begin{figure}
\includegraphics[width=\columnwidth]{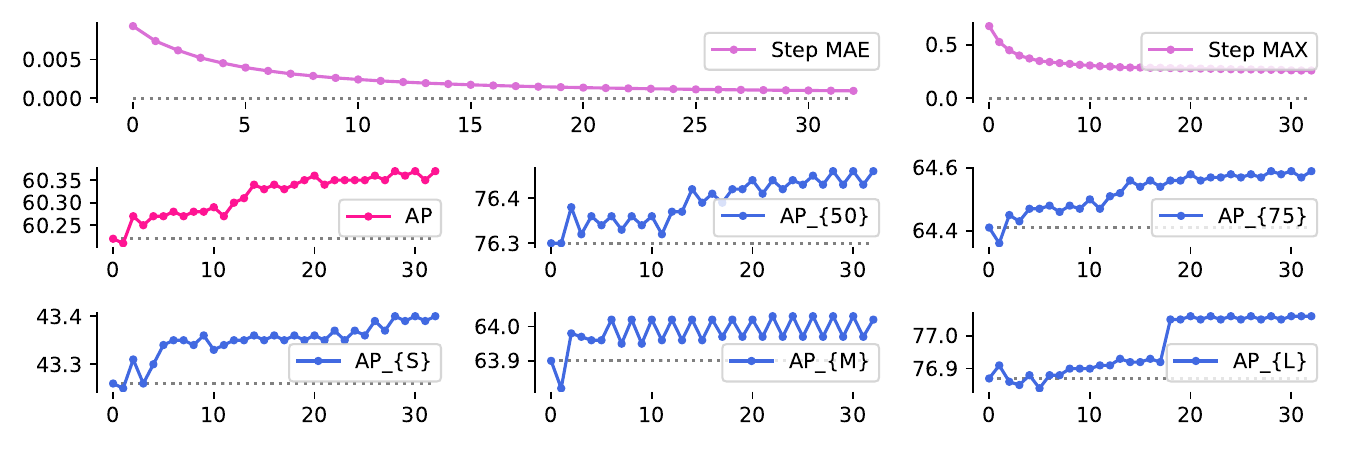}

\vspace{-0.7em}

\caption{Self Calibration Results with Subset Size 32.}

\label{fig:sc32}
\end{figure}

\begin{figure}
\includegraphics[width=\columnwidth]{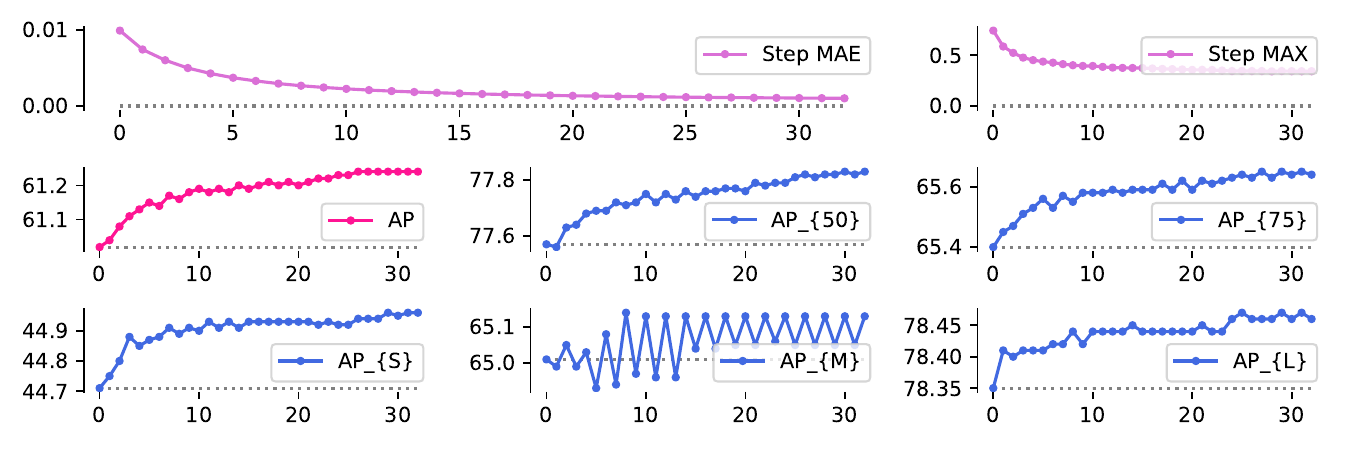}

\vspace{-0.7em}

\caption{Self Calibration Results with Subset Size 16.}

\label{fig:sc16}
\end{figure}

\begin{figure}
\includegraphics[width=\columnwidth]{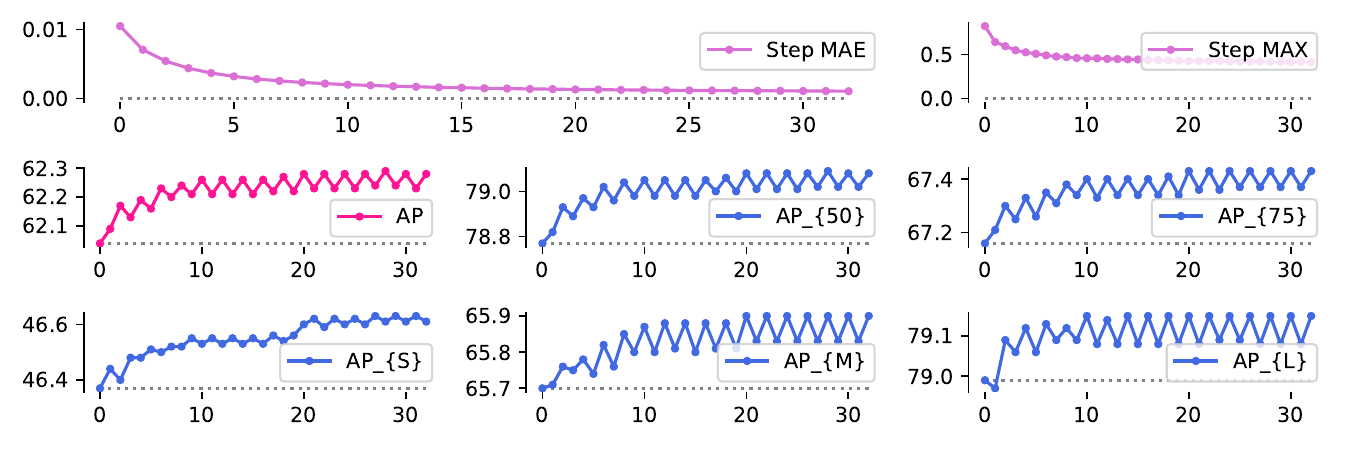}

\vspace{-0.7em}

\caption{Self Calibration Results with Subset Size 8.}

\label{fig:sc8}
\end{figure}

The self-calibration results for different subset sizes can be found
in Fig.~\ref{fig:sc256}, \ref{fig:sc128}, \ref{fig:sc64}, \ref{fig:sc32},
\ref{fig:sc16}, \ref{fig:sc8}. Besides the AP metrics, we track
two additional quantities: (1) ``Step MAE'': the step-wise mean
absolute error of $E_{c}$; (2) ``Step MAX'': the maximum absolute
value in the update step of $E_{c}$. The two values indicate whether
the algorithm is converging.

Take the subsets with size 256. The results are shown in Fig.~\ref{fig:sc256}.
An improvement in the overall AP can be seen along the self-calibration
process. The process is converging since the Step MAX will almost
decay to zero in the end.

For subsets of smaller sizes, the self-calibration process will also
lead to an improvement in the overall AP without any back-propagation,
as shown in Fig.~\ref{fig:sc128}, \ref{fig:sc64}, \ref{fig:sc32},
\ref{fig:sc16}, \ref{fig:sc8}. Differently, while the Step MAE is
always decreasing for any subset size, the Step MAX will saturate
at a larger value on smaller subsets. This is because smaller subsets
cannot provide a relatively stable statistics, and many conditional
probability values in $E_{c}$ will fluctuate during the self-calibration
process. Meanwhile, the AP curves will also fluctuate, especially
on the smallest subset, as shown in Fig.~\ref{fig:sc8}.

Convergence is important. When the model is deployed, it is not necessary
to self-calibrate on a fixed set of images, resulting in some inference
overhead. Instead, the method can use running statistics from the
past predictions to gradually calibrate itself towards the underlying
test distribution.

Based on these results, CaliDet can effectively calibrate itself with
its own predictions. In practice, more samples will make the process
more smooth in terms of AP curves.

\section{Ablation Studies \& Discussions}

In this section, we review the contribution of each component in the
proposed method, and discuss the limitations. In particular, CaliDet
comprises three main components in its implementation: (1) CaliFormer;
(2) Logit Manipulation loss, and (3) the edge sampling distribution
$\mathcal{E}$.

\textbf{CaliFormer} models the interaction among nodes $V$, and provides
the calibration vectors $V^{\prime}$. It is the only entry point
for prior injection as described in Eq.~(\ref{eq:califormer}). Thus,
replacing this component with standard Transformer will directly lead
to a not calibratable model. The scaling factor in Eq.~(\ref{eq:prediction-head})
is empirically set as $\rho=0.2$ for best performance. Setting it
to a larger value leads to a lower AP, while a smaller $\rho$ value
leads to a weaker effect with different deployment priors.

\textbf{Logit Manipulation.} This loss guides the model on the correct
usage of the information in the columns of $E$. Removing this loss
will lead to poor generalization and make the effect of calibration
almost vanish. Namely the model without LoMa does not satisfy the
expectation that the AP($E_{v}$) should be at least slightly greater
than AP($E_{t}$). This loss term has a hyper-parameter $\gamma$
to balance its magnitude with the other detection loss terms. Making
$\gamma$ larger means to encourage the model to follow the given
$E$ to a higher extent, but will meanwhile distract the model from
the detection loss and lead to a worse AP. We omit the corresponding
experimental results for brevity.

\textbf{Edge Sampling.} This provides the model with different $E$
during the training process for generalization to the arbitrary $E$
provided at run-time, as described in Sec.~\ref{sec:32}. For each
mini-batch, we randomly choose a prior from $\{E_{x},E_{b},E_{t}\}$,
namely the statistics for a single sample, the mini-batch, and the
whole training set. According to our observation, the diversity of
edges is very important for the model's generalization against different
priors. For instance, using only $E_{t}$ or $E_{x}$ leads to overfitting
at $E_{t}$ or $E_{x}$, and hence performs poorly at other different
edges.

Besides, a Gaussian noise is applied to the randomly chosen edge.
It increases the area covered by $\mathcal{E}$ during the training
process, and helps the model better generalize against different $E$.
For instance, the AP($E_{v}$) is lower than AP($E_{t}$) without
the noise, which is undesired. If its variance is excessively large,
the model will learn to discard the too noisy injection and end up
with a weak effect of calibration.

\begin{table}
\centering
\resizebox{1.0\linewidth}{!}{%
\setlength{\tabcolsep}{3pt}
\begin{tabular}{cccc}
\toprule 
Module & Backbone & Detection Transformer & CaliFormer\tabularnewline
\midrule 
Time (Ratio) & $17.10\pm2.88$ ($25.8\%$) & $47.86\pm7.06$ ($72.2\%$) & $1.37\pm0.30$ ($2.0\%$)\tabularnewline
\bottomrule
\end{tabular}}

\vspace{-0.5em}

\caption{Inference Time per Image. The unit is millisecond.}

\label{tab:time}
\end{table}

\textbf{Overhead.} The inference time cost with an NVIDIA RTX A5000
GPU is shown in Tab.~\ref{tab:time}. Compared to the backbone and
the detection transformer, our method only introduces a negligible
computational overhead through CaliFormer. In practice, since it is
not necessary to update the deployment prior upon every single forward
pass, the calibration vectors $V^{\prime}$ can be cached untill the
deployment prior update. In this way, the overhead introduced by our
method is further reduced based on the frequency of deployment prior
update.

\textbf{Limitations.} Deployment prior is the relation among different
object classes. For datasets with a small average number of objects
per image (e.g., Pascal VOC~\cite{pascal}), the performance gain
of prior injection could be marginal because the leverageable object
relation is relatively scarce.

\textbf{Future Work.} The core idea behind CaliDet can be interpreted
differently. If some statistical properties of the test distribution
can be characterized, we may be able to bind the model behavior with
such properties as an input, and calibrate the model at the run-time
for distribution shift.

\textbf{Supplementary} document contains elaborations on how to calculate
different edges, more technical details, experiments on D-DETR~\cite{deformabledetr},
as well as evaluations on Objects365~\cite{O365} dataset with the
model trained on COCO.

\section{Conclusions}

With a run-time calibratable object detector, the deployment prior
can be injected to adapt it to shifted object relation distribution
without any parameter update. The more accurate the injected prior
is, the better it performs. When no deployment prior is available,
the model can still leverage its own prediction results for self-calibration.

{\small

\bibliographystyle{ieeenat_fullname}
\bibliography{draft}

}

\newpage
\appendix

\section{More Technical Details \& Discussions}

\begin{figure}
\fbox{\includegraphics[width=\columnwidth]{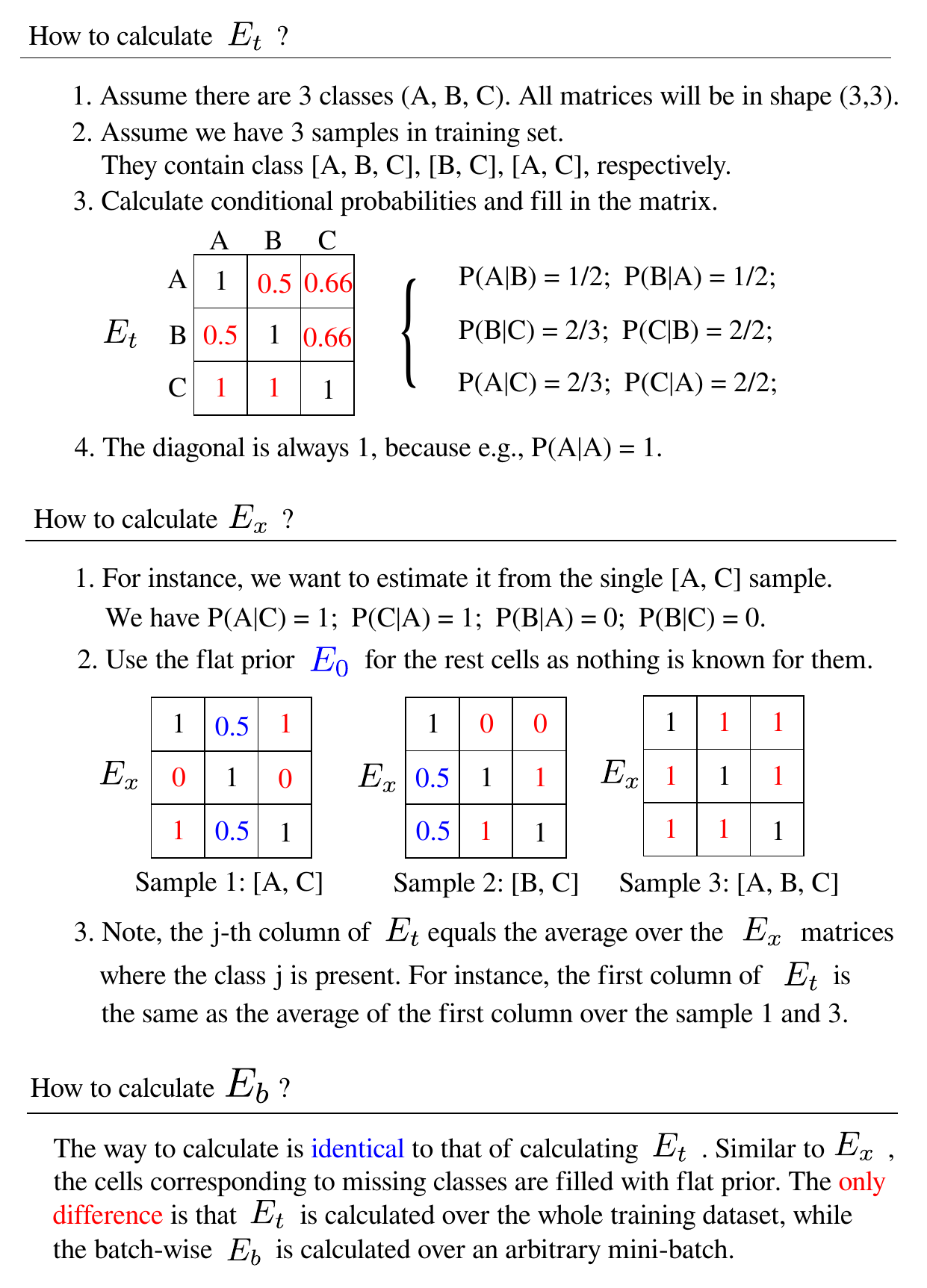}}

\vspace{-0.5em}

\caption{Detailed demonstration on the way to calculate $E_{t}$ and $E_{x}$.
The calculation method is also described in footnotes in the manuscript
where we introduce ``Binding Detector Behavior to $E$''.}

\label{fig:how-to-calculate-e}
\end{figure}

\subsection{Calculation of Matrix $E$}

\textbf{\uline{Definition.}} The edge $E\in[0,1]^{K\times K}$ is
a conditional probability matrix, where the $(i,j)$-th element is
$P(i|j)$, namely the probability that ``class $i$'' appears in
an image given ``the presence of class $j$''. This is already made
clear in the manuscript. To clearly demonstrate how the matrix $E$
is computed, we will provide examples in Fig.~\ref{fig:how-to-calculate-e},
and reference Python code implementation to make it crystal clear.

\textbf{(a). Flat Prior $E_{0}$.} Flat prior $E_{0}$ is the edge
when everything is completely unknown. Assume we have $K$ object
classes, and $0$ examples for calculating the $K\times K$ conditional
probabilities. In this case, as nothing is known about the class $i$
and class $j$, given the presence of class $j$, the probability
of the presence of class $i$ equals the probability of the absence
of class $i$. Thus both probabilities are $0.5$. Only when $i=j$,
the $P(i|j)$ will be $1$. Thus, as described in the footnotes in
the manuscript, $E_{0}$ is a symmetric matrix, where its diagonal
values are $1$, and all the off-diagonal values are $0.5$:
\[
E_{0}=\begin{bmatrix}1 & 0.5 & 0.5 & \cdots & 0.5\\
0.5 & 1 & \ddots & \ddots & \vdots\\
0.5 & \ddots & \ddots & \ddots & 0.5\\
\vdots & \ddots & \ddots & 1 & 0.5\\
0.5 & \cdots & 0.5 & 0.5 & 1
\end{bmatrix}.
\]

Based on the flat prior, if the probability $P(i|j)<0.5$, it can
be interpreted as ``class $i$ is less likely to appear given the
presence of class $j$'', and there is already some information about
class $i$ and $j$. Likewise, if the probability $P(i|j)>0.5$, it
can be interpreted as ``class $i$ is more likely to appear given
the presence of class $j$''.

\textbf{\uline{Example Python code for \mbox{$E_{0}$}:}}
\begin{lstlisting}
import numpy as np

def edge_flat_prior(num_cls: int) -> np.ndarray:
    E = np.ones((num_cls, num_cls)) * 0.5
    np.fill_diagonal(E, 1.0)
    return E

print(edge_flat_prior(5))
\end{lstlisting}

\textbf{(b). Calculation of $E_{t}$, $E_{v}$, $E_{b}$, and $E_{x}$.}
The only difference among them is the number of samples used to calculate
the statistics (conditional probability). Specifically, $E_{t}$ is
calculated on the whole COCO training set. The $E_{v}$ is calculated
on the whole COCO validation set. The $E_{b}$ is dynamically calculated
on an arbitrary given mini-batch during the training or inference
stage. The $E_{x}$ is dynamically calculated for a single image during
the training or inference stage. The notation $E_{b(2)}$, for instance,
means the $E_{b}$ calculated on a two-image mini-batch. By definition,
$E_{t}$ is equivalent to $E_{b}$ by treating the whole training
dataset as a mini-batch. Note, following the idea of flat prior $E_{0}$,
when the class $j$ does not exist in the population at all, we use
the corresponding value in the flat prior.

Recall that $P(i|j)=P(i,j)/P(j)$. Therefore, the calculation is to
count the co-occurrence of $(i,j)$ classes, and then divide it by
the occurrence count of class $j$. See Fig.~\ref{fig:how-to-calculate-e}
for detailed examples. 

We also use $\overline{E_{x}}$ in the manuscript, which is a ``flipped''
version of $E_{x}$. This provides a ``misleading'' deployment prior
for testing our model as a sanity check for whether the model behavior
follows the input $E$. The calculation is identical to $E_{x}$,
except for that we fill $1$ in the matrix where $P(i|j)=0$, and
$0$ where $P(i|j)=1$, and the the diagonal is left intact.

\textbf{\uline{Example Python code for \mbox{$E_{x}$}:}}

\begin{lstlisting}
from typing import *
import numpy as np

sample1 = [0,2]  # contains A, C
sample2 = [1,2]  # contains B, C
sample3 = [0,1,2]  # contains A, B, C

def edge_sample(num_cls: int,
                labels: List[int]) -> np.ndarray:
    E = np.ones((num_cls, num_cls)) * 0.5
    for j in labels:
        E[:, j] = 0.0
        E[labels, j] = 1.0
    np.fill_diagonal(E, 1.0)
    return E

print(edge_sample(3, sample1))
print(edge_sample(3, sample2))
print(edge_sample(3, sample3))
\end{lstlisting}

\textbf{\uline{Example Python code for \mbox{$E_{b}$}:}}

\begin{lstlisting}
from typing import *
import numpy as np
import itertools as it

sample1 = [0,2]  # contains A, C
sample2 = [1,2]  # contains B, C
sample3 = [0,1,2]  # contains A, B, C

def sparse2dense(num_cls: int,
                 targets: List[List[int]],
                 ) -> np.ndarray:
    dense = np.zeros((num_cls, len(targets)))
    for (j, labels) in enumerate(targets):
        dense[labels, j] = 1.0
    return dense

def dense2edge(dense: np.array) -> np.ndarray:
    num_cls, batch_size = dense.shape
    E = np.ones((num_cls, num_cls)) * 0.5
    rN = range(num_cls)
    for (i, j) in it.product(rN, rN):
        cij = np.logical_and(dense[i,:],
              dense[j,:]).sum()
        cj = dense[j,:].sum()
        if cj > 0:
            pij = cij / cj
        else:
            pij = 0.5  # flat prior
        E[i,j] = pij
    np.fill_diagonal(E, 1.0)
    return E

def edge_batch(num_cls: int,
               targets: List[List[int]],
               ) -> np.ndarray:
    dense = sparse2dense(num_cls, targets)
    E = dense2edge(dense)
    return E

batch = [sample1, sample2, sample3]
print(edge_batch(3, batch))
\end{lstlisting}

The Python code for $E_{t}$ is completely identical to $E_{b}$ by
treating the whole COCO training set as a single batch. Note, the
above Python code snippets are only for demonstration. Our actual
implementation is optimized using Cython.

\subsection{Regarding Graph $\langle V,E\rangle$}

\textbf{(a). Conditional Probability.} The extent that class $j$
depends on class $i$ does not necessarily equal the extent that the
class $i$ depends on class $j$. The edge should be directional (asymmetric).
Since conditional probability is asymmetric, it can indicate the relation
in a directional way. In contrast, symmetric relations, such as co-occurrence
probability and correlation are not suitable here.

\textbf{(b). Implicitly Learned Edge.} Learning the edges implicitly,
without a clear definition for its entries, could make the interpretation
of model behavior and self-calibration difficult. In this paper, we
use manually selected conditional probability for describing the characteristics
of distribution and its shift for interpretability. Implicitly learned
distribution characteristics are left for future study.

\textbf{(c). Columns of $E$.} The $j$-th column are the conditional
probabilities that the class $i$ appears given the presence of $j$.
If $P(i|j)=1.0$, it means class $i$ always appears together with
$j$. As a result, the model learns the representation of class $j$
with the information from class $i$ entangled. If $P(i|j)=0.0$,
the representation of class $j$ will suppress the information from
class $i$. The $j$-th column vector of $E$ represents exactly how
each class $i$ contributes to the representation of class $j$. Using
the $j$-th row for class $j$ is incorrect by design.

\begin{figure}
\includegraphics[width=\linewidth]{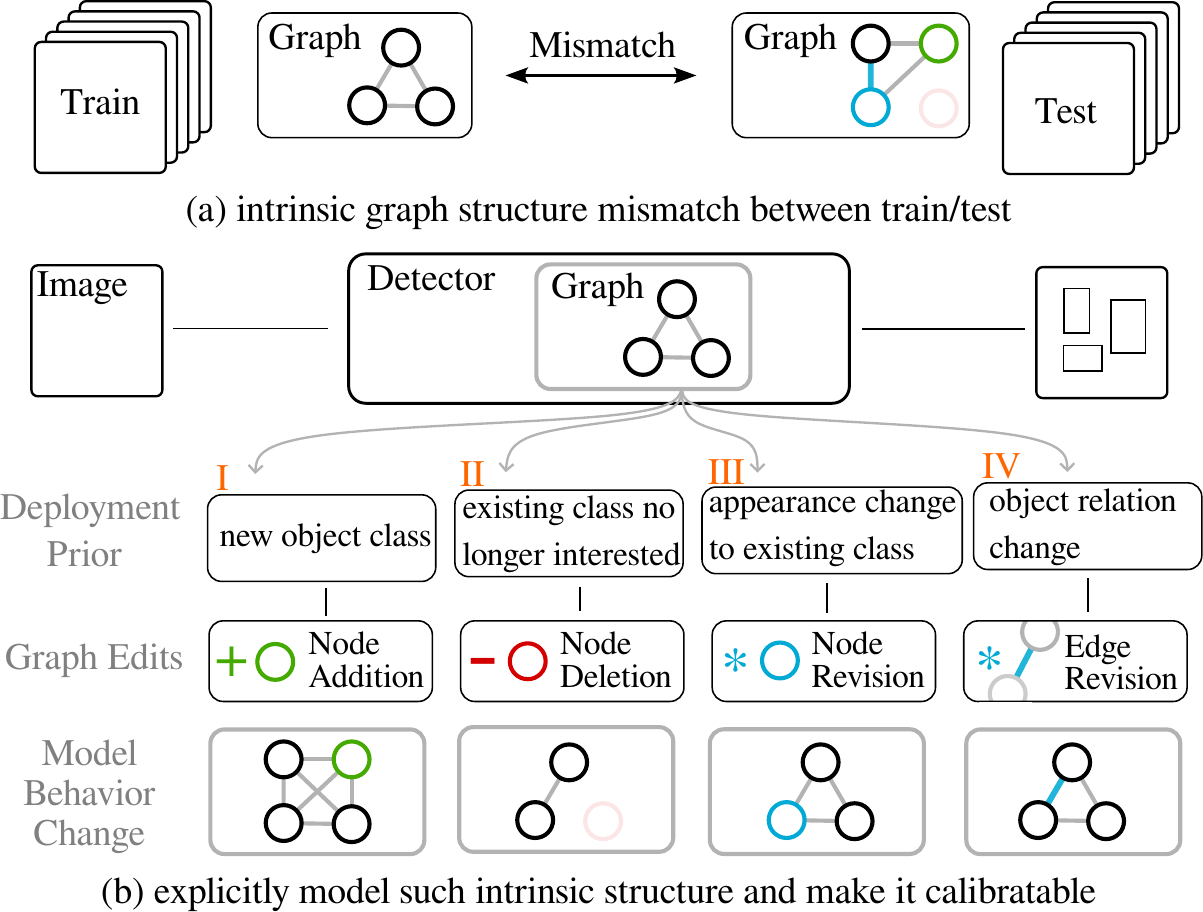}

\caption{Calibration of both nodes $V$ and edges $E$ is a more complicated
and challenging issue. The node addition, node deletion, and node
revision are conceptually incremental learning or transfer learning.
Our current model only implements the fourth type of deployment prior
injection, namely the edge revision, because this is uncharted in
the literature. The other three types of graph edits are widely explored
for object detection. Combining all of these types of graph edits
could be one of the future research directions.}

\label{fig:calidet-withnode}
\end{figure}

\textbf{(d). Calibration of nodes $V$.} As explained in the manuscript,
the calibration of nodes $V$ is conceptually incremental or transfer
learning, which is beyond the scope of this paper. See Fig.~\ref{fig:calidet-withnode}
for the demonstration. We leave this as a direction for future study.

\textbf{(e). Underlying Idea.} The core idea of CaliDet can be isolated
from the object detection task. Assuming that some statistical properties
of the test distribution can be characterized, we may be able to bind
the model behavior with such properties as an input, and calibrate
the model at the run-time for the distribution shift without any change
to the model parameter. This problem setting is even more challenging
than test-time adaptation, but more efficient in terms of computation
cost. In particular, the ``statistical property'' used in this paper
is object relation, namely the conditional probability between two
object classes. If we regard such object relation as second-order
information, we can consider using different statistical properties
in other tasks. For instance, in long-tail image classification, we
may leverage the class distribution, which is first-order information.
We leave these possibilities for future study.

\subsection{Model Design and Training}

\subsubsection{CaliFormer and Prediction Head}

\textbf{(a). Model Capacity.} The default number of layers of CaliFormer
is 3. When the number of layers is decreased from 3 to 1, the capacity
of CaliFormer is insufficient for the model to generalize against
different edges will decrease. When the number of layers is increased
to more than 3, the performance gain through injection turns to be
marginal.

\textbf{(b). $\Delta E$ and flat prior $E_{0}$.} The difference
$\Delta E$ is designed to represent the extent of change for the
dependency of one class to another class, compared to the case where
nothing is known. We should not calculate $\Delta E$ as $E-E_{t}$,
because $E_{t}$ is already biased towards the training set. Only
the flat prior can present ``nothing is known''.

\textbf{(c). Scaling Factor $\rho$.} This hyper-parameter should
not be too large. A large $\rho$ such as $1.0$ makes the classification
layer excessively distracted to harm the classification accuracy.
We empirically set it as $0.2$ based on our observations.

\textbf{(d). Decoder Layers.} The detector involves 6 layers of Decoders.
Each layer of the decoder has its own classification heads and an
auxiliary loss. We share the same CaliFormer among all the prediction
heads for all 6 decoder layers.

\textbf{(e). Compatibility.} CaliDet is not coupled with any particular
DETR-like detector, and can be incorporated into other DETR-like models.
It only contains three core components: (1) calibration vector; (2)
LoMa loss; (3) edge sampling distribution. This document includes
results of CaliDet based on D-DETR~\cite{deformabledetr} instead
of DINO~\cite{dino}. We leave the one-stage and two-stage detectors
for future study.

\subsubsection{Logit Matching Loss (LoMA)}

\begin{figure*}
\includegraphics[width=\linewidth]{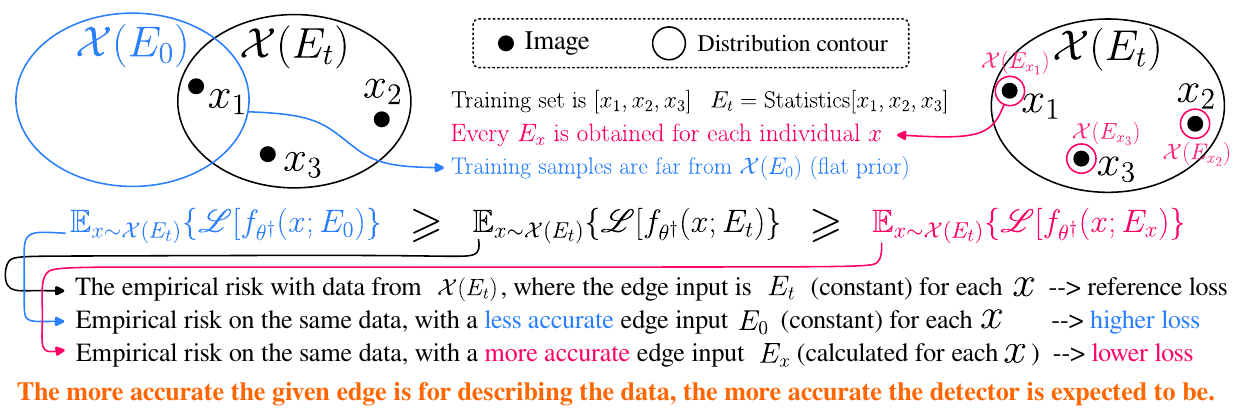}

\caption{Illustration of Eq. 5 in the manuscript.}

\label{fig:eq5}
\end{figure*}

The LoMa loss term is proposed in order to make the loss landscape
vary based on the values of the input matrix $E$. Without it, the
loss landscape is invariant to the input $E$, and cannot properly
and effectively guide the model towards the correct usage of information
in $E$. It is designed to indirectly enforce Eq. 5 in the manuscript.
The illustration of Eq. 5 can be found in Fig.~\ref{fig:eq5}.

Without the LoMA loss function, the model fails to generalize and
satisfy the expected order of AP given different $E$. This is because
the model lacks an explicit penalty to learn to leverage the information
in $E$. The loMA loss weight $\gamma$ can be adjusted. We empirically
set it as $20$ for both DINO and D-DETR variants of CaliDet for the
best results.

\subsubsection{Edge Sampling Distribution}

\textbf{(a). When no $\{E_{t}\}$ in $\mathcal{E}$.} If the $E_{t}$
is removed, the model will not generalize well around $E_{t}$, as
AP($E_{t}$) is lower than AP($E_{0}$), failing the sanity test on
the expected AP order.

\textbf{(b). When only $\{E_{x}$\} in $\mathcal{E}$.} This leads
to very poor generalization around $E_{t}$. Namely the AP($E_{t}$)
is much lower than AP($E_{0}$), which fails the sanity test. Diversity
is needed in the sampling distribution. Results are omitted.

\textbf{(c). When no Gaussian noise in $\mathcal{E}$. }This leads
to poor generalization and fails to satisfy the expected order of
AP. Namely, the AP($E_{v}$) is slightly lower than AP($E_{t}$),
which indicates the failure of generalization against different edges
despite their differences. Noise is necessary.

\textbf{(d). Magnitude of Gaussian Noise.} A slightly smaller Gaussian
noise leads to a larger gap between AP($E_{t}$) and AP($E_{0}$),
but a smaller gap between AP($E_{t}$) and AP($E_{v}$). A too small
Gaussian noise will result in overfitting to the constant edges such
as $E_{t}$, and fail to generalize against different priors. In contrast,
if the Gaussian noise is too large, the model will gradually lose
the sensitivity against the injected prior, due to the information
being too noisy. We empirically select the Gaussian parameter as $\mathcal{N}(0,0.16^{2})$
for best generalization. 

\textbf{(e).} \textbf{Parameter Tuning.} One important clue to observe
during hyper-parameter tuning is to make sure that $E_{v}$ performance
is slightly higher than $E_{t}$ -- only in this way can the model
correctly generalize against new conditional probability shifts. When
adapting CaliDet to a new model with a different training batch size,
the other edges such as $E_{b(4)}$ (statistics on 4-sample mini-batch)
can be added to the edge sampling algorithm to increase edge diversity
during training. For instance, this is done in the CaliDet (D-DETR)
experiments in this supplementary document.

\subsection{Inference \& Self-Calibration}

\textbf{(a). Run-time Calibration.} During the inference stage, the
model architecture, as well as the parameters of CaliDet is completely
frozen. The back-propagation is not allowed in the ``run-time calibration''
setting. See Section.~\ref{subsec:Advantage-of-Run-Time} for the
discussion on the advantages of a run-time calibration. The calibration
(i.e., deployment prior injection) process is done by simply changing
the input matrix $E$. Note, our calibratable detector accepts an
additional input $E$ besides the image.

\textbf{(b). Cached $V^{\prime}$.} If the input $E$ is a fixed constant,
we can cache the calculated calibration vector $V'=g(V,\Delta E)$,
and reduce the CaliFormer overhead to nearly zero, depending on the
update frequency of $E$. In this case the only computational overhead
is the matrix addition operation: $W+\rho V'$. In practice, the deployment
prior does not have to be changed upon every forward pass. Instead,
the deployment prior can be updated periodically, for instance once
per hour. Thus, as long as $E$ does not frequently change, the computation
cost of run-time calibration can be reduced to nearly zero by such
an engineering technique to cache the calibration vectors. And the
inference cost will solely depend on the baseline detector itself
when the deployment prior is not changed.

\textbf{(c). Initialization with $E_{t}$.} Our self-calibration process
is initialized as $E_{c}\leftarrow E_{t}$, inheriting the training
set prior. This means that we assume the test distribution is close
to the training set when nothing is known about the test set at the
beginning of self-calibration. If the algorithm is initialized with
$E_{0}$, the AP will be lower than AP($E_{t}$). Thus, we empirically
initialize $E_{c}$ from the default prior $E_{t}$.

\subsubsection{Advantage of Run-Time Calibration}

\label{subsec:Advantage-of-Run-Time}

\textbf{(1).} If a calibratable detector is used on hundreds of edge
devices, they can be calibrated toward their local distribution shifts.
The calibration process is merely changing the $K\times K$ input
matrix $E$ without any change to the network architecture or the
parameters. That means the calibration process can be done by weak
devices locally. Compared to existing works that adapt a detector
and involve back-propagation, fine-tuning a detector for hundreds
of devices with some new data is still not a negligible cost, which
is repetitive and much more costly than run-time calibration.

\textbf{(2).} Through the self-calibration process, CaliDet can calibrate
itself towards the test data distribution without human annotation.
Since the default AP($E_{t}$) of CaliDet is on par with the baseline
detector, it is usable as a drop-in replacement for a general detector.
Thus, designing a run-time calibratable detector is very challenging.

The core underlying idea of CaliDet is in fact independent to object
detection. Assume a training dataset has some physically interpretable
bias that can be described by a set of parameters. Presumably, a deep
model well-learned on such a dataset will also inherit such bias.
Such bias can be either harmful or helpful depending on the proximity
of the test distribution and the training distribution. If we supplement
the physically meaningful parameters directly to the model input,
and let the model understand what these parameters mean, we should
be able to adapt the model towards the distribution shift with respect
to the mentioned parameters by merely changing the input parameters.

This work can also be seen as extending the object detector to be
promptable, where the format of such ``prompt'' is fixed as the
conditional probability matrix $E$. We hope this work could inspire
readers beyond the area of object detection. For instance, can we
consider other types of human-understandable training set characteristics,
and design more controllable deep network models with run-time calibration
capability?

\subsection{Visualization of Self-Calibration}

\begin{figure}
\includegraphics[width=\linewidth]{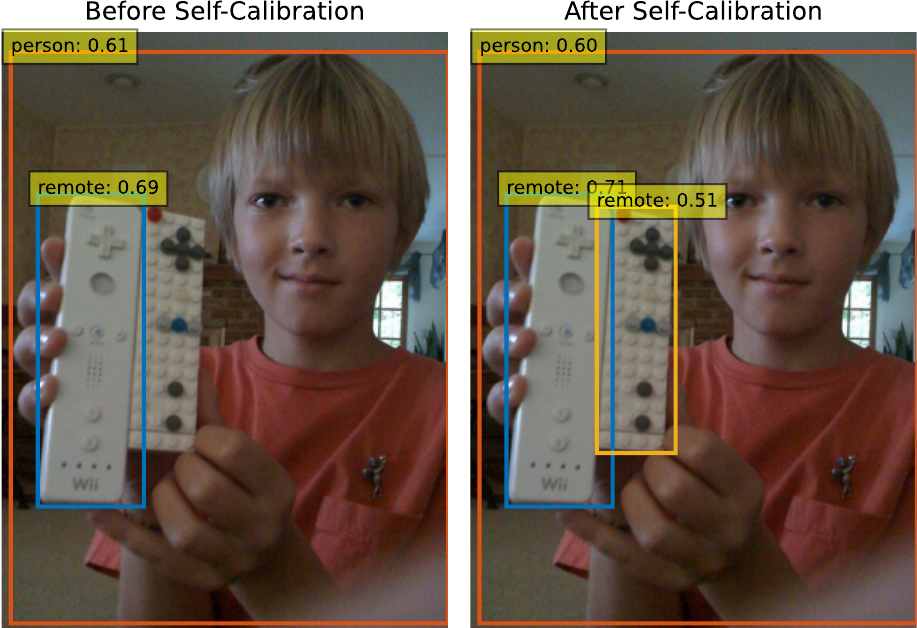}

\includegraphics[width=\linewidth]{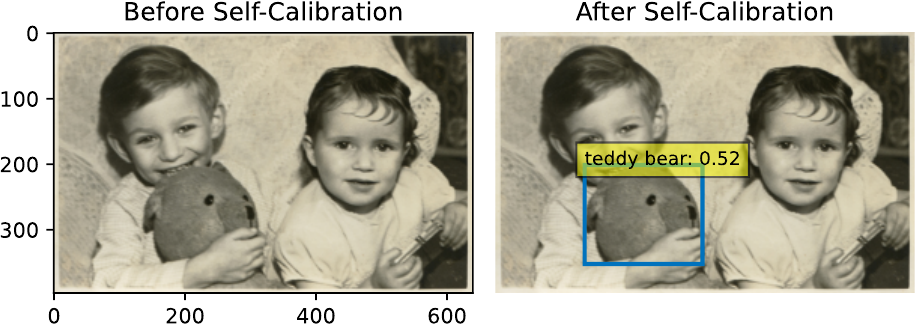}

\includegraphics[width=\linewidth]{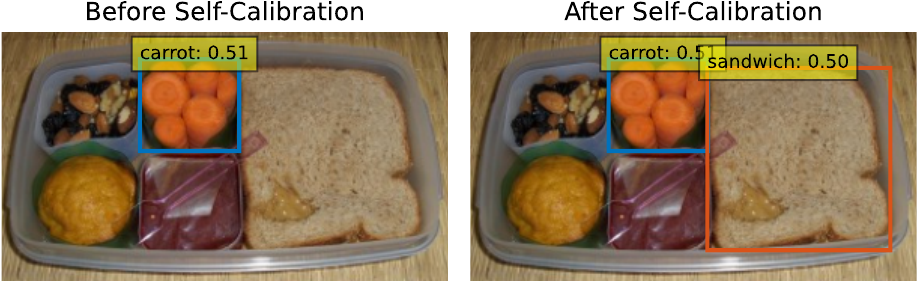}

\includegraphics[width=\linewidth]{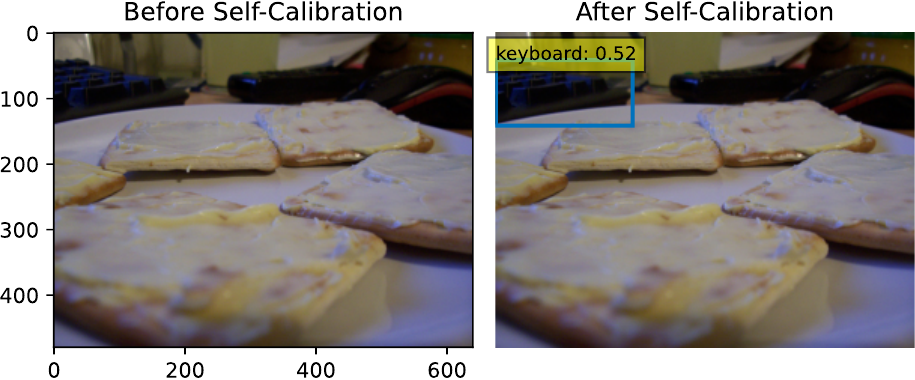}

\caption{Visualization of Self-Calibration with CaliDet (DINO). The corresponding
subset size is 256. (Part 1 of 2)}

\label{fig:dino-vis-256-1}
\end{figure}

\begin{figure}
\includegraphics[width=\linewidth]{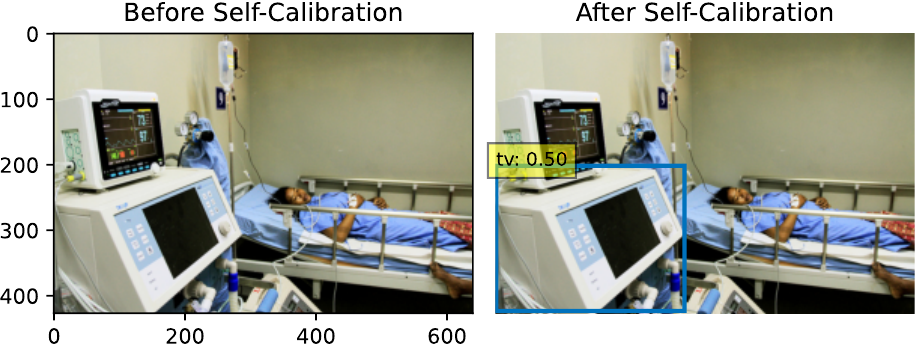}

\includegraphics[width=\linewidth]{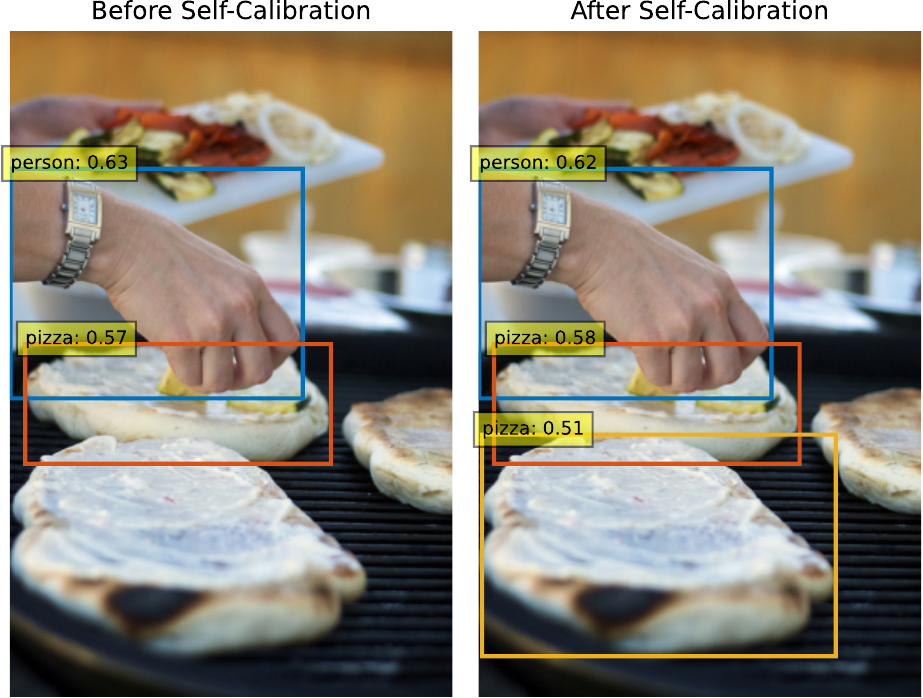}

\includegraphics[width=\linewidth]{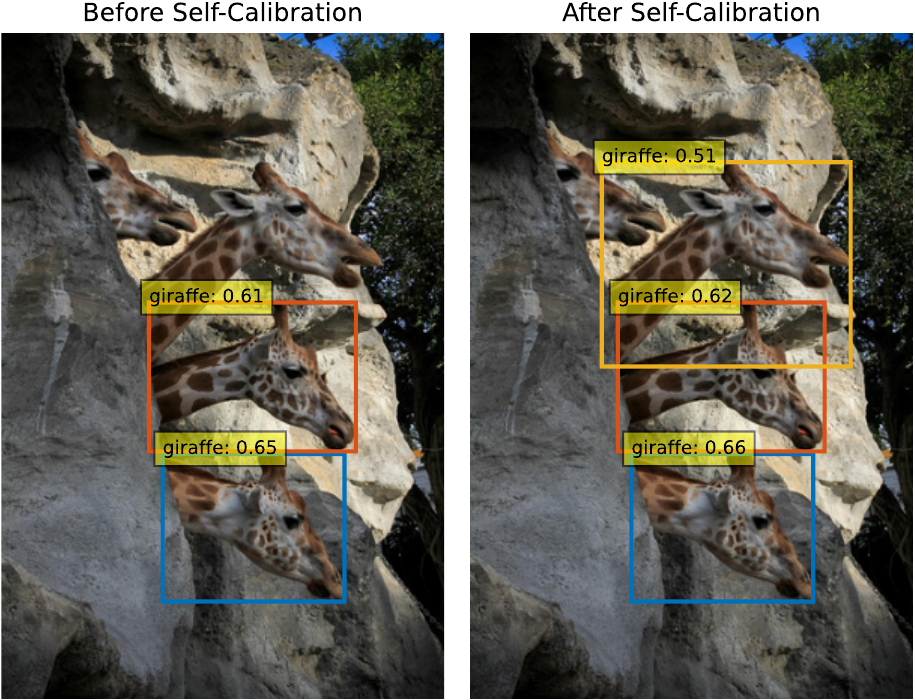}

\includegraphics[width=\linewidth]{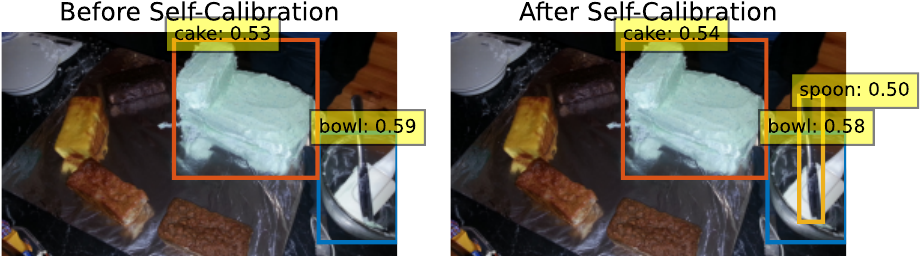}

\caption{Visualization of Self-Calibration with CaliDet (DINO). The corresponding
subset size is 256. (Part 2 of 2)}

\label{fig:dino-vis-256-2}
\end{figure}

\begin{figure}
\includegraphics[width=\linewidth]{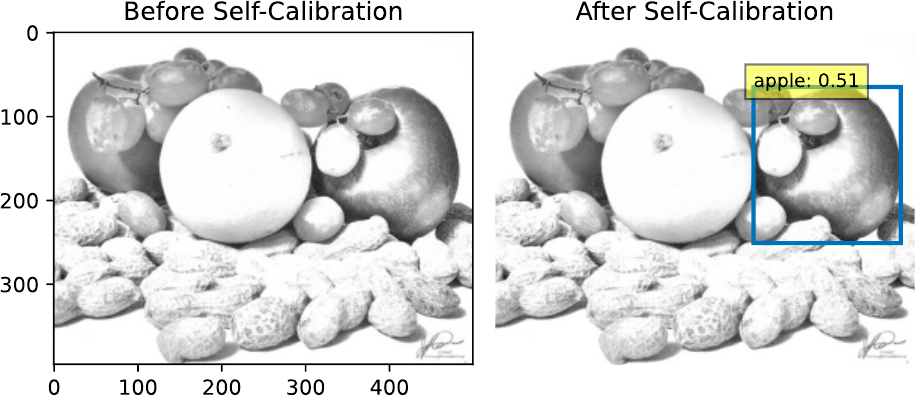}

\includegraphics[width=\linewidth]{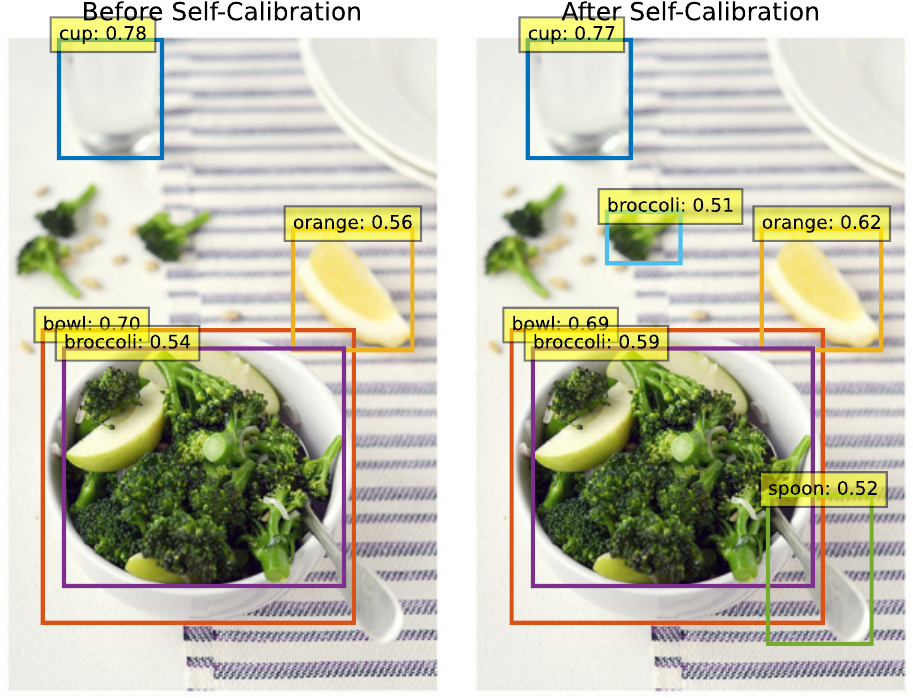}

\caption{Visualization of Self-Calibration with CaliDet (DINO). The corresponding
subset size is 8.}

\label{fig:dino-vis-8}
\end{figure}

We provide some visualization results for the self-calibration algorithm.
We filter out the detection results with a score less than $0.5$.
For self-calibration results on subsets of size 256, some results
can be found in Fig.~\ref{fig:dino-vis-256-1} and Fig.~\ref{fig:dino-vis-256-2}.
For self-calibration results on subsets of size 8, some results can
be found in Fig.~\ref{fig:dino-vis-8}. As can be seen from the figures,
after deployment prior injection or self-calibration, the model's
confidence will increase, and hence, some missed detections will appear
after calibration. Self-calibration is effective.

\section{Porting to D-DETR~\cite{deformabledetr}}

\begin{table}
\resizebox{1.0\linewidth}{!}{%
\setlength{\tabcolsep}{4pt}%
\begin{tabular}{cccccccc}
\toprule 
\multirow{2}{*}{\textbf{Method}} & \multirow{2}{*}{\textbf{Injection}} & \multicolumn{6}{c}{\textbf{Standard COCO Metrics}}\tabularnewline
\cmidrule{3-8}
 &  & AP & AP$_{50}$ & AP$_{75}$ & AP$_{S}$ & AP$_{M}$ & AP$_{L}$\tabularnewline
\midrule
\midrule 
D-DETR & - & 41.47 & 61.87 & 44.73 & 24.18 & 45.30 & 55.95\tabularnewline
\midrule 
\rowcolor{blue!7}\cellcolor{white} & $\overline{E_{x}}$ & 36.38 & 51.95 & 39.77 & 18.53 & 39.92 & 51.72\tabularnewline
 & $E_{0}$ & 41.49 & 60.78 & 44.96 & 23.52 & 45.41 & 56.17\tabularnewline
\rowcolor{black!7}\cellcolor{white}CaliDet & $E_{t}$ & 41.69 & 61.82 & 45.02 & 23.52 & 45.42 & 56.39\tabularnewline
(D-DETR) & $E_{v}$ & 41.70 & 61.84 & 45.03 & 23.52 & 45.43 & 56.41\tabularnewline
 & $E_{b}$ & 43.66 & 65.17 & 46.69 & 25.59 & 47.34 & 58.15\tabularnewline
\rowcolor{magenta!7}\cellcolor{white} & $E_{x}$ & 43.92 & 65.61 & 47.22 & 25.68 & 47.55 & 58.85\tabularnewline
\bottomrule
\end{tabular}

}
\vspace{0.1em}

\caption{Standard COCO Evaluation Results. We incorporate our proposed CaliDet
method to the D-DETR~\cite{deformabledetr} model (single scale variant).
The results suggest that our method is effective on D-DETR, as the
performance order given different deployment priors is as expected.}

\label{tab:ddetr-const}
\end{table}

\begin{table}
\resizebox{1.0\linewidth}{!}{%
\setlength{\tabcolsep}{3pt}%

\begin{tabular}{ccccccccc}
\toprule 
\multirow{2}{*}{\textbf{Subset Size}} & \multirow{2}{*}{\textbf{Injection}} & \multirow{2}{*}{$\epsilon$} & \multicolumn{6}{c}{\textbf{COCO Metrics Averaged over Subsets}}\tabularnewline
\cmidrule{4-9}
 &  &  & \multicolumn{1}{c}{\cellcolor{red!7}AP} & \multicolumn{1}{c}{AP$_{50}$} & \multicolumn{1}{c}{\cellcolor{red!7}AP$_{75}$} & \multicolumn{1}{c}{AP$_{S}$} & \multicolumn{1}{c}{\cellcolor{red!7}AP$_{M}$} & \multicolumn{1}{c}{AP$_{L}$}\tabularnewline
\midrule
\midrule 
\multirow{3}{*}{8} & \cellcolor{black!7}$E_{t}$ & 0 & 55.35 & 75.34 & 59.72 & 34.91 & 58.79 & 73.96\tabularnewline
 & $E_{b(8)}$ & 0.331 & 55.64 & 75.80 & 60.00 & 35.18 & 59.02 & 74.32\tabularnewline
 &  &  & {\footnotesize\textcolor{magenta}{(+ 0.29)}} & {\footnotesize\textcolor{magenta}{(+ 0.46)}} & {\footnotesize\textcolor{magenta}{(+ 0.28)}} & {\footnotesize\textcolor{magenta}{(+ 0.27)}} & {\footnotesize\textcolor{magenta}{(+ 0.23)}} & {\footnotesize\textcolor{magenta}{(+ 0.36)}}\tabularnewline
\midrule
\multirow{3}{*}{16} & \cellcolor{black!7}$E_{t}$ & 0 & 53.64 & 73.31 & 57.80 & 33.50 & 57.23 & 72.35\tabularnewline
 & $E_{b(16)}$ & 0.276 & 53.85 & 73.66 & 57.98 & 33.65 & 57.36 & 72.56\tabularnewline
 &  &  & {\footnotesize\textcolor{magenta}{(+ 0.21)}} & {\footnotesize\textcolor{magenta}{(+ 0.32)}} & {\footnotesize\textcolor{magenta}{(+ 0.18)}} & {\footnotesize\textcolor{magenta}{(+ 0.15)}} & {\footnotesize\textcolor{magenta}{(+ 0.13)}} & {\footnotesize\textcolor{magenta}{(+ 0.21)}}\tabularnewline
\midrule
\multirow{3}{*}{32} & \cellcolor{black!7}$E_{t}$ & 0 & 52.30 & 71.61 & 56.38 & 32.16 & 55.76 & 70.75\tabularnewline
 & $E_{b(32)}$ & 0.202 & 52.48 & 71.85 & 56.56 & 32.31 & 55.88 & 70.91\tabularnewline
 &  &  & {\footnotesize\textcolor{magenta}{(+ 0.18)}} & {\footnotesize\textcolor{magenta}{(+ 0.24)}} & {\footnotesize\textcolor{magenta}{(+ 0.18)}} & {\footnotesize\textcolor{magenta}{(+ 0.15)}} & {\footnotesize\textcolor{magenta}{(+ 0.12)}} & {\footnotesize\textcolor{magenta}{(+ 0.16)}}\tabularnewline
\midrule
\multirow{3}{*}{64} & \cellcolor{black!7}$E_{t}$ & 0 & 50.54 & 69.76 & 54.54 & 31.02 & 54.51 & 68.71\tabularnewline
 & $E_{b(64)}$ & 0.122 & 50.70 & 69.99 & 54.69 & 31.08 & 54.55 & 68.83\tabularnewline
 &  &  & {\footnotesize\textcolor{magenta}{(+ 0.16)}} & {\footnotesize\textcolor{magenta}{(+ 0.23)}} & {\footnotesize\textcolor{magenta}{(+ 0.15)}} & {\footnotesize\textcolor{magenta}{(+ 0.06)}} & {\footnotesize\textcolor{magenta}{(+ 0.04)}} & {\footnotesize\textcolor{magenta}{(+ 0.12)}}\tabularnewline
\midrule
\multirow{3}{*}{128} & \cellcolor{black!7}$E_{t}$ & 0 & 48.36 & 67.57 & 52.18 & 29.97 & 52.75 & 66.40\tabularnewline
 & $E_{b(128)}$ & 0.062 & 48.48 & 67.71 & 52.31 & 30.03 & 52.77 & 66.49\tabularnewline
 &  &  & {\footnotesize\textcolor{magenta}{(+ 0.12)}} & {\footnotesize\textcolor{magenta}{(+ 0.14)}} & {\footnotesize\textcolor{magenta}{(+ 0.13)}} & {\footnotesize\textcolor{magenta}{(+ 0.06)}} & {\footnotesize\textcolor{magenta}{(+ 0.02)}} & {\footnotesize\textcolor{magenta}{(+ 0.09)}}\tabularnewline
\midrule
\multirow{3}{*}{256} & \cellcolor{black!7}$E_{t}$ & 0 & 46.21 & 65.62 & 49.77 & 28.78 & 50.60 & 63.93\tabularnewline
 & $E_{b(256)}$ & 0.034 & 46.31 & 65.76 & 49.86 & 28.78 & 50.64 & 63.95\tabularnewline
 &  &  & {\footnotesize\textcolor{magenta}{(+ 0.10)}} & {\footnotesize\textcolor{magenta}{(+ 0.14)}} & {\footnotesize\textcolor{magenta}{(+ 0.09)}} & {\footnotesize\textcolor{magenta}{(+ 0.00)}} & {\footnotesize\textcolor{magenta}{(+ 0.04)}} & {\footnotesize\textcolor{magenta}{(+ 0.02)}}\tabularnewline
\bottomrule
\end{tabular}

}
\vspace{0.1em}

\caption{Subset Evaluation with Varying Subset Size. This experiment is based
on D-DETR and COCO dataset. We split all validation data to subsets
with different sizes. This means with a frozen, we can still obtain
some AP improvements for free (\ie, without any gradient update)
given an arbitrary distribution shift in terms of the conditional
probability.}

\label{tab:ddetr-mpi}
\end{table}

\begin{figure}[h]
\includegraphics[width=\linewidth]{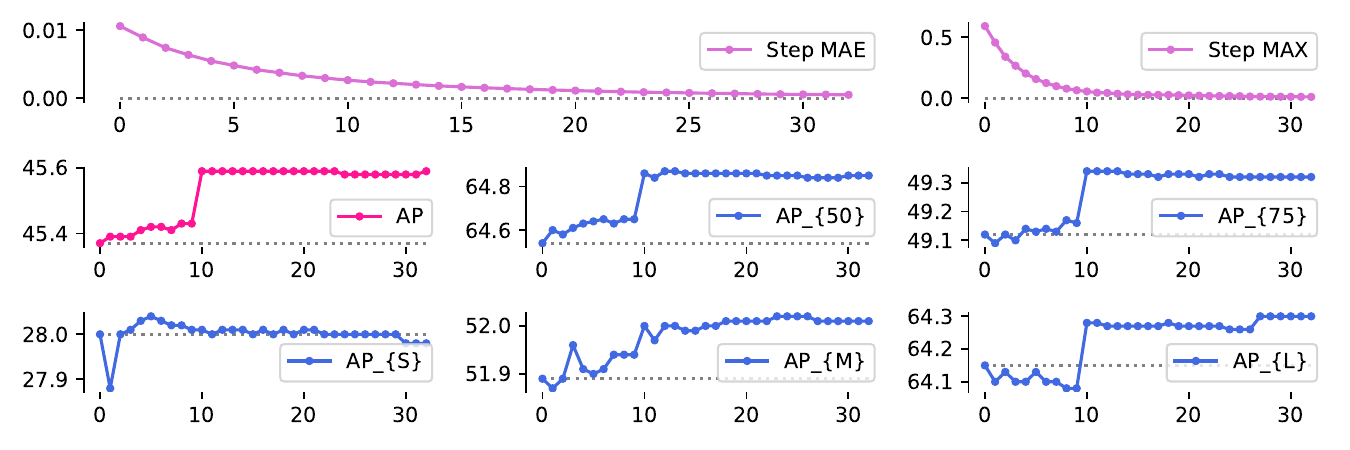}

\caption{Self-calibration results on COCO dataset using CaliDet (D-DETR) w/
subset size 256.}

\label{fig:qe-5-256}
\end{figure}

\begin{figure}[h]
\includegraphics[width=\linewidth]{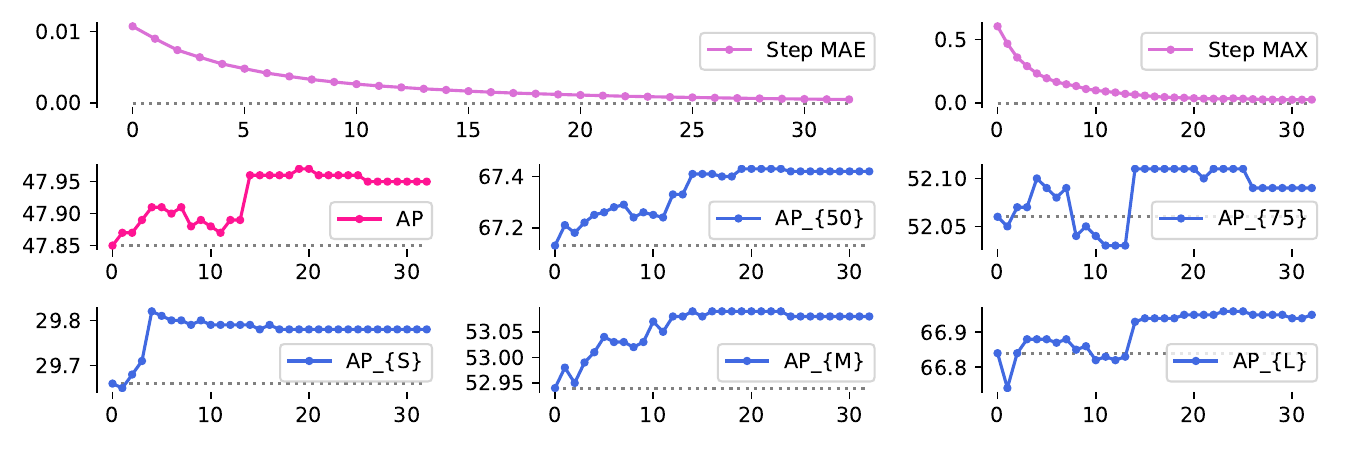}

\caption{Self-calibration results on COCO dataset using CaliDet (D-DETR) w/
subset size 128.}

\label{fig:qe-5-128}
\end{figure}

\begin{figure}[h]
\includegraphics[width=\linewidth]{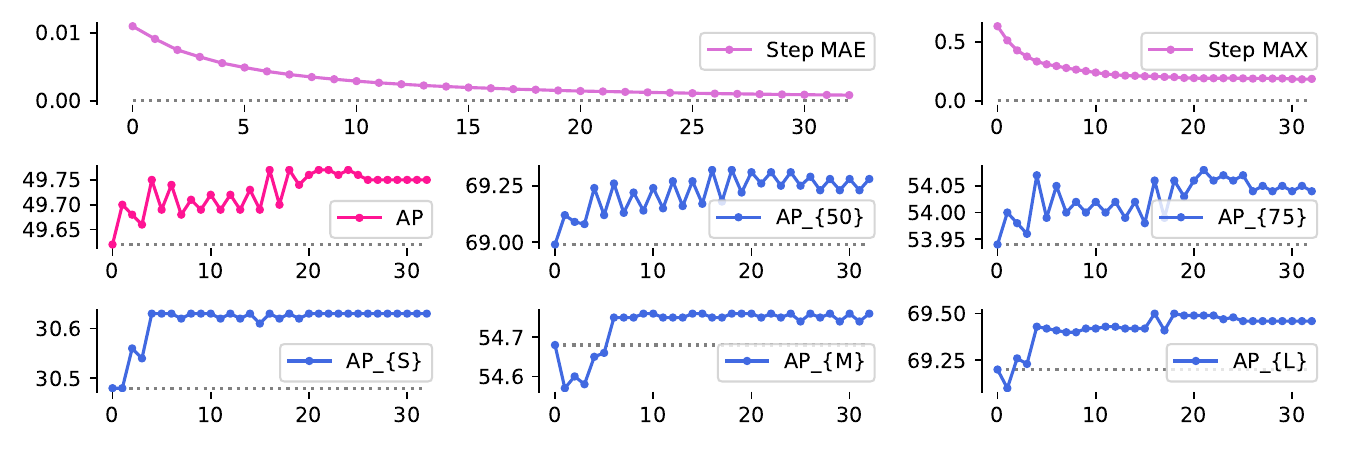}

\caption{Self-calibration results on COCO dataset using CaliDet (D-DETR) w/
subset size 64.}

\label{fig:qe-5-64}
\end{figure}

\begin{figure}[h]
\includegraphics[width=\linewidth]{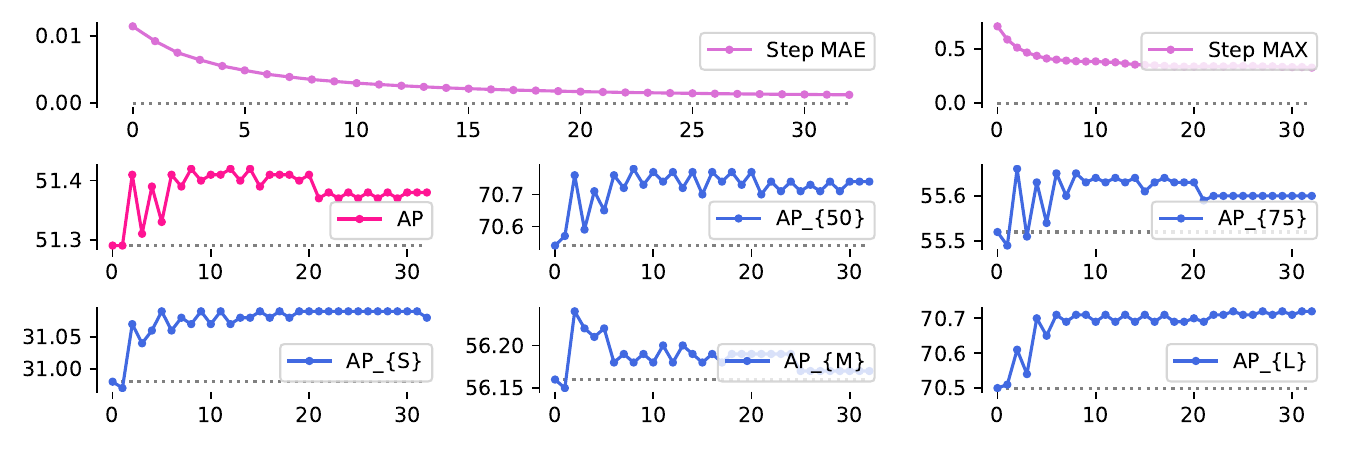}

\caption{Self-calibration results on COCO dataset using CaliDet (D-DETR) w/
subset size 32.}

\label{fig:qe-5-32}
\end{figure}

\begin{figure}[h]
\includegraphics[width=\linewidth]{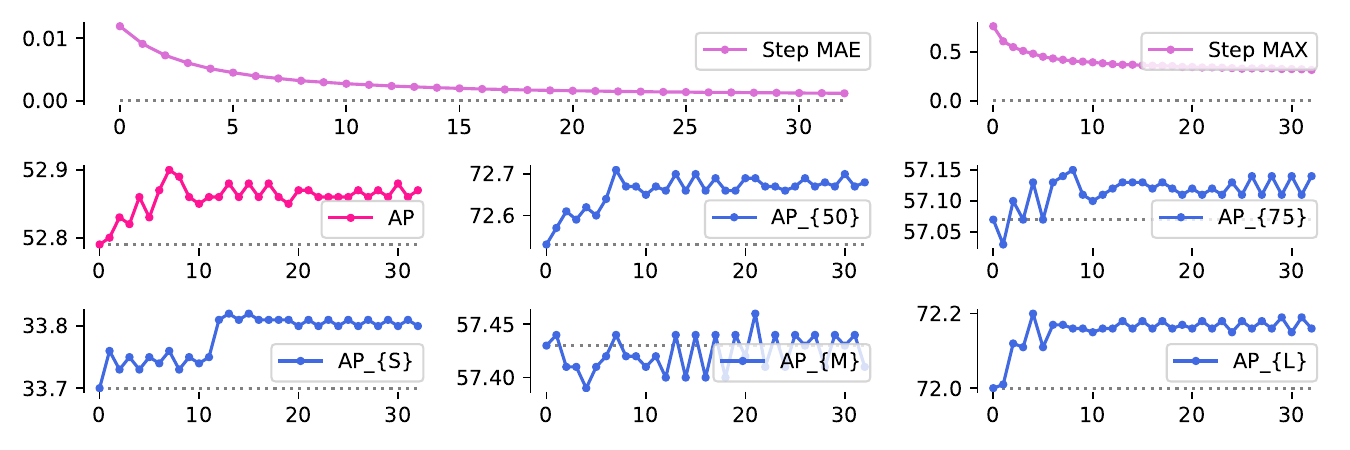}

\caption{Self-calibration results on COCO dataset using CaliDet (D-DETR) w/
subset size 16.}

\label{fig:qe-5-16}
\end{figure}

\begin{figure}[h]
\includegraphics[width=\linewidth]{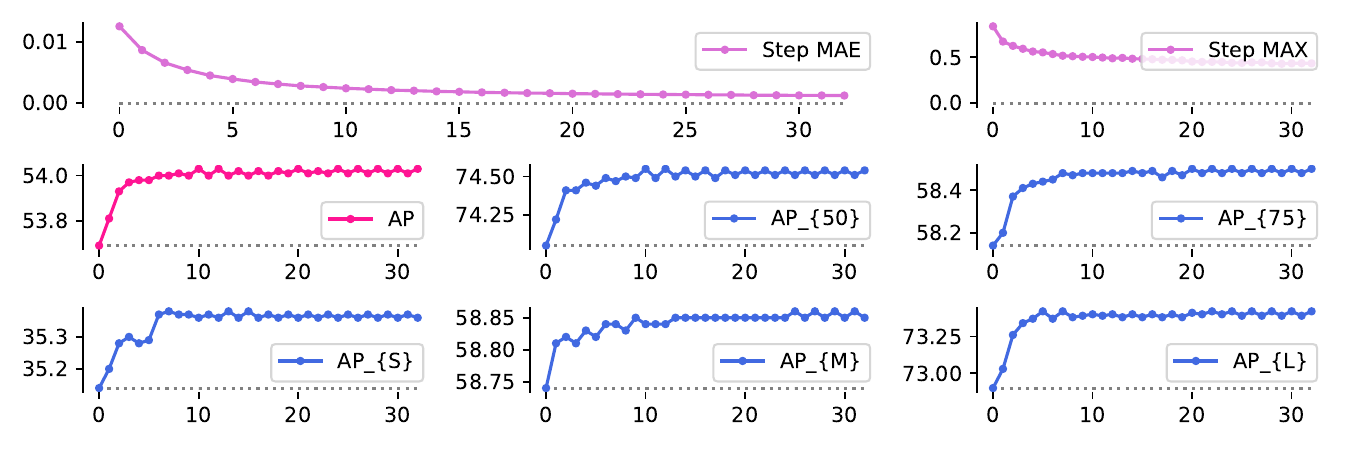}

\caption{Self-calibration results on COCO dataset using CaliDet (D-DETR) w/
subset size 8.}

\label{fig:qe-5-8}
\end{figure}

In this subsection, we implement our proposed method onto Deformable-DETR~\cite{deformabledetr}
(abbr., D-DETR), which is also commonly referred to as one of the
state-of-the-art method in the recent detection literature besides
DINO~\cite{dino}. We adopt the single-scale variant of D-DETR as
the base model. According to the following experiments, our method
is effective on D-DETR.

\subsection{Experimental Setup and Results}

There are three parts of evaluations. (1) The standard COCO evaluation
results can be found in Tab.~\ref{tab:ddetr-const}. (2) The subset
evaluations can be found in Tab.~\ref{tab:ddetr-mpi}. (3) The self-calibration
results for varying subset sizes can be found in Fig.~\ref{fig:qe-5-256},
Fig.~\ref{fig:qe-5-128}, Fig.~\ref{fig:qe-5-64}, Fig.~\ref{fig:qe-5-32},
Fig.~\ref{fig:qe-5-16}, and Fig.~\ref{fig:qe-5-8}. All these results
show that CaliDet is effective on the D-DETR model. We conclude that
CaliDet is not coupled with DINO, and we speculate that CaliDet should
be effective on other DETR-like models.

\subsection{Visualization of Self-Calibration}

\begin{figure}
\includegraphics[width=\linewidth]{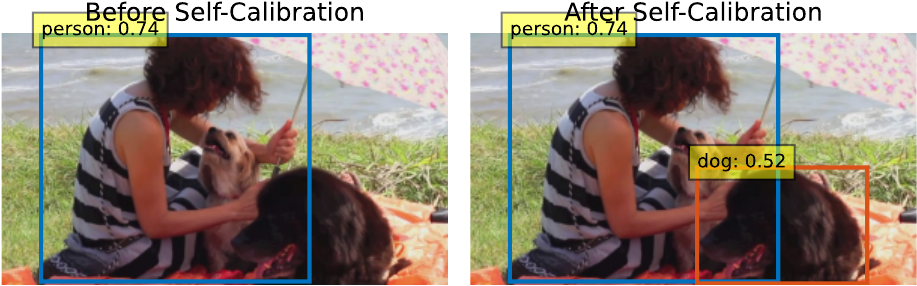}

\includegraphics[width=\linewidth]{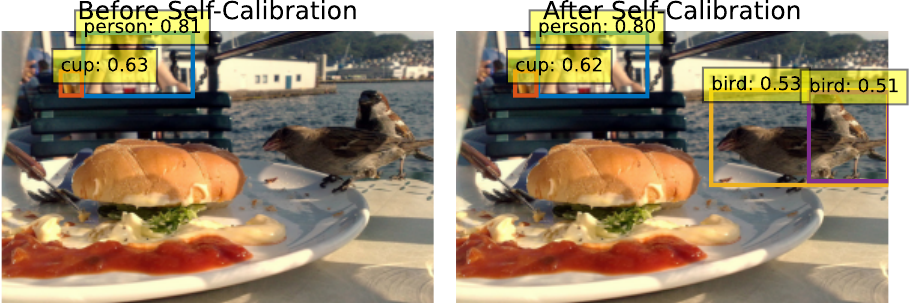}

\includegraphics[width=\linewidth]{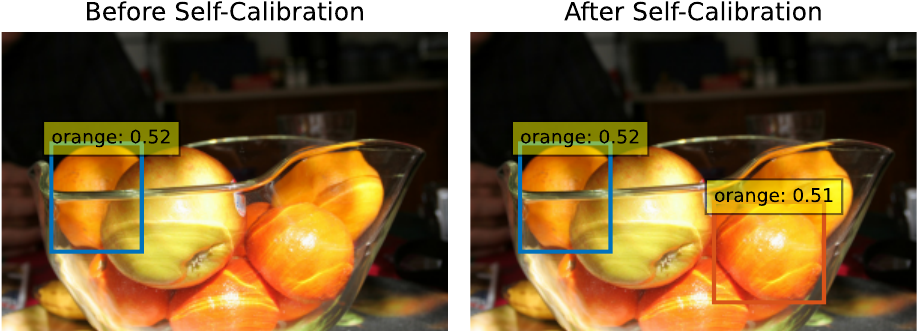}

\includegraphics[width=\linewidth]{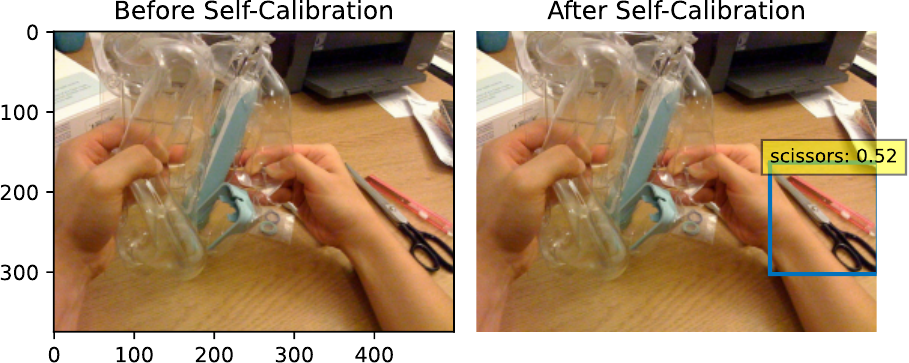}

\includegraphics[width=\linewidth]{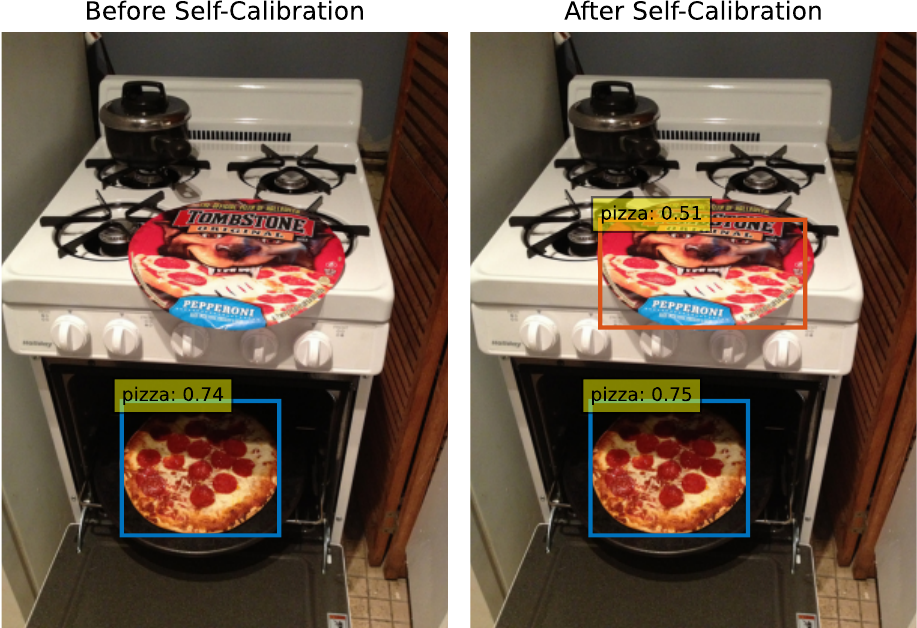}

\caption{Visualization of Self-Calibration with CaliDet (D-DETR). The corresponding
subset size is 256.}

\label{fig:ddetr-vis-256}
\end{figure}

We provide some visualization results for the self-calibration algorithm.
We filter out the detection results with a score of less than $0.5$.
For self-calibration results on subsets of size 256, some results
can be found in Fig.~\ref{fig:ddetr-vis-256}. As can be seen from
the figures, after deployment prior injection or self-calibration,
the model's confidence will increase, and hence some missed detections
will appear after calibration.

\section{Objects365 Evaluation}

Objects365~\cite{O365} (O365) from ICCV2019 is a dataset designed
to spur object detection research with a focus on diverse objects
in the Wild. It involves 365 categories, 2 million images, and 30
million bounding boxes, which is more challenging than the COCO 2017
detection dataset.

In this section, we first freeze the CaliDet trained on the COCO training
set (it has not seen any O365 training sample). Then we directly evaluate
CaliDet on the Objects365 validation set without any change in the
model architecture, or the parameters. Namely, the original classification
heads are kept and used without any fine-tuning.

Similar to the COCO evaluations in the manuscript, we will show the
standard evaluations on the O365 validation set, the O365 validation
subset evaluation, as well as self-calibration on O365 subsets.

\subsection{Dataset Preprocessing}

\begin{table}
\resizebox{\linewidth}{!}{%
\begin{tabular}{cc}
\toprule 
COCO Class (in JSON format) & Objects365 Class (in JSON format)\tabularnewline
\midrule
\midrule 
\multirow{3}{*}{\texttt{\{'id': 16, 'name': 'bird'\}}} & \texttt{\{'name': 'Wild Bird', 'id': 57\}}\tabularnewline
\cmidrule{2-2}
 & \texttt{\{'name': 'Pigeon', 'id': 165\}}\tabularnewline
\cmidrule{2-2}
 & \texttt{\{'name': 'Parrot', 'id': 320\}}\tabularnewline
\midrule 
\texttt{\{'id': 31, 'name': 'handbag'\}} & \texttt{\{'name': 'Handbag/Satchel', 'id': 14\}}\tabularnewline
\midrule 
\texttt{\{'id': 33, 'name': 'suitcase'\}} & \texttt{\{'name': 'Luggage', 'id': 121\}}\tabularnewline
\midrule 
\texttt{\{'id': 35, 'name': 'skis'\}} & \texttt{\{'name': 'Skiboard', 'id': 120\}}\tabularnewline
\midrule 
\multirow{10}{*}{\texttt{\{'id': 37, 'name': 'sports ball'\}}} & \texttt{\{'name': 'Soccer', 'id': 119\}}\tabularnewline
\cmidrule{2-2}
 & \texttt{\{'name': 'Other Balls', 'id': 157\}}\tabularnewline
\cmidrule{2-2}
 & \texttt{\{'name': 'Baseball', 'id': 166\}}\tabularnewline
\cmidrule{2-2}
 & \texttt{\{'name': 'Basketball', 'id': 178\}}\tabularnewline
\cmidrule{2-2}
 & \texttt{\{'name': 'Billards', 'id': 190\}}\tabularnewline
\cmidrule{2-2}
 & \texttt{\{'name': 'American Football', 'id': 207\}}\tabularnewline
\cmidrule{2-2}
 & \texttt{ddd\{'name': 'Tennis', 'id': 211\}}\tabularnewline
\cmidrule{2-2}
 & \texttt{\{'name': 'Volleyball', 'id': 240\}}\tabularnewline
\cmidrule{2-2}
 & \texttt{\{'name': 'Golf Ball', 'id': 248\}}\tabularnewline
\cmidrule{2-2}
 & \texttt{\{'name': 'Table Tennis ', 'id': 365\}}\tabularnewline
\midrule 
\texttt{\{'id': 51, 'name': 'bowl'\}} & \texttt{\{'name': 'Bowl/Basin', 'id': 27\}}\tabularnewline
\midrule 
\texttt{\{'id': 55, 'name': 'orange'\}} & \texttt{\{'name': 'Orange/Tangerine', 'id': 105\}}\tabularnewline
\midrule 
\texttt{\{'id': 67, 'name': 'dining table'\}} & \texttt{\{'name': 'Dinning Table', 'id': 99\}}\tabularnewline
\midrule 
\texttt{\{'id': 72, 'name': 'tv'\}} & \texttt{\{'name': 'Moniter/TV', 'id': 38\}}\tabularnewline
\midrule 
\texttt{\{'id': 88, 'name': 'teddy bear'\}} & \texttt{\{'name': 'Stuffed Toy', 'id': 71\}}\tabularnewline
\midrule 
\texttt{\{'id': 89, 'name': 'hair drier'\}} & \texttt{\{'name': 'Hair Dryer', 'id': 329\}}\tabularnewline
\bottomrule
\end{tabular}}

\caption{Class Mapping between COCO Dataset and Objects365 Dataset. We build
this mapping in order to directly evaluate the CaliDet (trained solely
on COCO training set) on the O365 validation set. The mapping for
unmentioned classes can be automatically built using case-insensitive
matching in category names. Here are some details regarding the mapping:
(1) We manually confirmed the ``suitcase'' (COCO) and the ``Luggage''
(O365) correspondence by examples; (2) Most ``sports balls'' in
COCO are baseball and tennis; (3) ``Dinning Table'' is a typo in
O365; (4) O365 contains more types of stuffed toys than ``Teddy bear''.}

\label{tab:coco-o365-map}
\end{table}

The COCO involves 80 object categories, while O365 involves 365 categories.
Before the experiments, we first trim the O365 validation set and
remap the classes to COCO classes, because it is impossible to use
a frozen COCO model to detect the non-overlapping categories in O365.

We first manually build a class mapping between COCO and O365. Most
of the 80 categories can match with a class in O365 with case-insensitive
matching with their names. For all mismatching classes, we manually
check example images, and build the mappings as shown in Tab.~\ref{tab:coco-o365-map}.

Then we filter the O365 annotations that do not associate with any
COCO class, as well as the O365 images that contain zero COCO object
category. The original O365 validation dataset contains $80,000$
images, and $1,240,587$ annotations. After removing the non-COCO
classes, $71,316$ images ($89.1\%$) and $438,215$ ($35.2\%$) annotations
remain.

\begin{figure}
\includegraphics[width=\columnwidth]{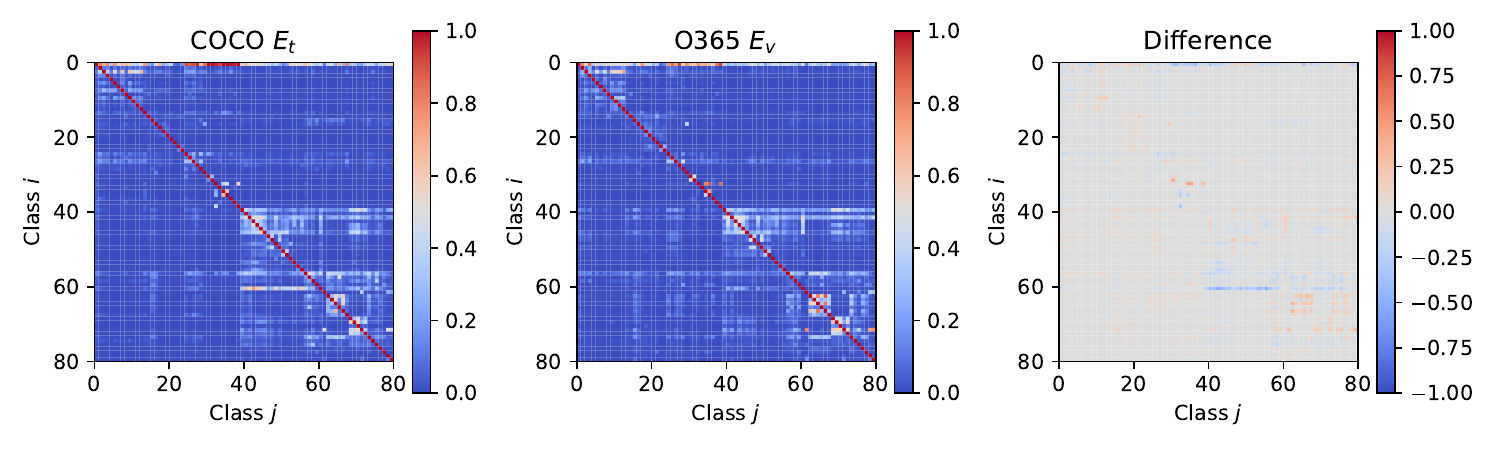}

\caption{Comparison between $E_{t}^{\text{COCO}}$ and $E_{v}^{\text{O365}}$.
The MAE between them is $0.018$. The max and min absolute differences
are $0.526$ and $0.0$, respectively. The 50-, 90-, 97-th percentile
values are 0.004, 0.048, and 0.122, respectively.}

\label{fig:cet2oev}
\end{figure}

\begin{figure}
\includegraphics[width=\columnwidth]{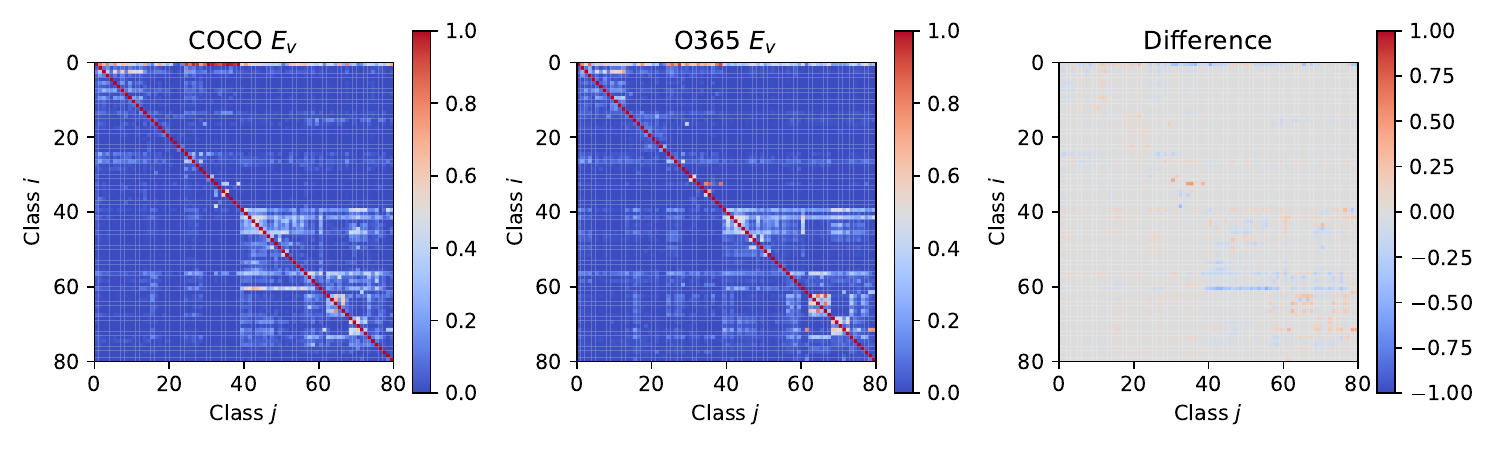}

\caption{Comparison between $E_{v}^{\text{COCO}}$ and $E_{v}^{\text{O365}}$.
The MAE between them is $0.020$. The max and min absolute differences
are $0.506$ and $0.0$, respectively. The 50-, 90-, 97-th percentile
values are 0.005, 0.054, and 0.132, respectively. Note, the COCO validation
statistics $E_{v}^{\text{COCO}}$ deviates from $E_{v}^{O365}$ further
than the COCO training set statistics $E_{t}^{\text{COCO}}$. Thus,
the AP on O365 when we inject $E_{v}^{\text{COCO}}$ should be slightly
lower than $E_{t}^{\text{COCO}}$, because the former one has a larger
discrepancy compared to the ground-truth $E_{v}^{O365}$.}

\label{fig:cev2oev}
\end{figure}

In the end, we map the object categories in the remaining data and
annotations back to the COCO category IDs. The evaluations in this
section are based on this processed O365 validation set. The O365
validation set conditional probability $E_{v}^{\text{O365}}$, as
well as its difference with respect to $E_{v}^{\text{COCO}}$ and
$E_{t}^{\text{COCO}}$ can be found in Fig.~\ref{fig:cet2oev} and
Fig.~\ref{fig:cev2oev}. Note, the COCO validation statistics $E_{v}^{\text{COCO}}$
deviates from $E_{v}^{O365}$ further than the COCO training set statistics
$E_{t}^{\text{COCO}}$. Thus, the AP on O365 when we inject $E_{v}^{\text{COCO}}$
should be slightly lower than $E_{t}^{\text{COCO}}$, because the
former one has a larger discrepancy compared to the ground truth $E_{v}^{O365}$.

\subsection{Experimental Results}

\begin{table}
\resizebox{1.0\linewidth}{!}{%
\setlength{\tabcolsep}{5pt}%
\begin{tabular}{cccccccc}
\toprule 
\multirow{2}{*}{\textbf{Method}} & \multirow{2}{*}{\textbf{Injection}} & \multicolumn{6}{c}{\textbf{Standard Metrics on O365}}\tabularnewline
\cmidrule{3-8}
 &  & AP & AP$_{50}$ & AP$_{75}$ & AP$_{S}$ & AP$_{M}$ & AP$_{L}$\tabularnewline
\midrule
\midrule 
\rowcolor{blue!7}\cellcolor{white} & $\overline{E_{x}}$ & 24.17 & 31.16 & 26.17 & 7.64 & 20.17 & 38.33\tabularnewline
 & $E_{0}$ & 32.80 & 43.15 & 35.51 & 12.98 & 29.50 & 47.01\tabularnewline
\rowcolor{black!7}\cellcolor{white}CaliDet & $E_{t}^{\text{COCO}}$ & 33.10 & 43.72 & 35.82 & 13.16 & 29.78 & 47.06\tabularnewline
(DINO) & $E_{v}^{\text{COCO}}$ & 33.08 & 43.69 & 35.80 & 13.16 & 29.76 & 47.03\tabularnewline
\rowcolor{yellow}\cellcolor{white} & $E_{v}^{\text{O365}}$ & 33.15 & 43.79 & 35.88 & 13.20 & 29.82 & 47.11\tabularnewline
 & $E_{b}$ & 38.80 & 51.82 & 41.98 & 16.94 & 35.69 & 52.83\tabularnewline
\rowcolor{magenta!7}\cellcolor{white} & $E_{x}$ & 39.18 & 52.34 & 42.40 & 17.29 & 36.05 & 53.24\tabularnewline
\bottomrule
\end{tabular}

}
\vspace{0.1em}

\caption{Standard Detection Metrics on Objects365 (O365) Dataset. MAE($E_{t}^{\text{COCO}}$,
$E_{v}^{\text{O365}}$)=$0.018$. MAE($E_{v}^{\text{COCO}}$, $E_{v}^{\text{O365}}$)=$0.020$.
Our model is effective. The AP on O365 when we inject $E_{v}^{\text{COCO}}$
is slightly lower than $E_{t}^{\text{COCO}}$, because the former
one has a larger discrepancy compared to the ground-truth $E_{v}^{O365}$
(marked in yellow).}

\label{tab:qc-1-o3}
\end{table}

\begin{table}
\resizebox{1.0\linewidth}{!}{%
\setlength{\tabcolsep}{3pt}%

\begin{tabular}{ccccccccc}
\toprule 
\multirow{2}{*}{\textbf{Subset Size}} & \multirow{2}{*}{\textbf{Injection}} & \multirow{2}{*}{$\epsilon$} & \multicolumn{6}{c}{\textbf{Metrics Averaged over O365 Subsets}}\tabularnewline
\cmidrule{4-9}
 &  &  & \multicolumn{1}{c}{\cellcolor{red!7}AP} & \multicolumn{1}{c}{AP$_{50}$} & \multicolumn{1}{c}{\cellcolor{red!7}AP$_{75}$} & \multicolumn{1}{c}{AP$_{S}$} & \multicolumn{1}{c}{\cellcolor{red!7}AP$_{M}$} & \multicolumn{1}{c}{AP$_{L}$}\tabularnewline
\midrule
\midrule 
\multirow{3}{*}{8} & \cellcolor{black!7}$E_{t}$ & 0 & 55.27 & 68.56 & 58.93 & 32.94 & 54.94 & 72.39\tabularnewline
 & $E_{b(8)}$ & $0.336$ & 55.62 & 69.04 & 59.30 & 33.35 & 55.26 & 72.57\tabularnewline
 &  &  & {\footnotesize\textcolor{magenta}{(+ 0.35)}} & {\footnotesize\textcolor{magenta}{(+ 0.48)}} & {\footnotesize\textcolor{magenta}{(+ 0.37)}} & {\footnotesize\textcolor{magenta}{(+ 0.41)}} & {\footnotesize\textcolor{magenta}{(+ 0.32)}} & {\footnotesize\textcolor{magenta}{(+ 0.18)}}\tabularnewline
\midrule
\multirow{3}{*}{16} & \cellcolor{black!7}$E_{t}$ & 0 & 52.29 & 65.08 & 55.79 & 32.80 & 52.39 & 69.61\tabularnewline
 & $E_{b(16)}$ & $0.292$ & 52.67 & 65.56 & 56.19 & 33.13 & 52.70 & 69.82\tabularnewline
 &  &  & {\footnotesize\textcolor{magenta}{(+ 0.38)}} & {\footnotesize\textcolor{magenta}{(+ 0.48)}} & {\footnotesize\textcolor{magenta}{(+ 0.40)}} & {\footnotesize\textcolor{magenta}{(+ 0.33)}} & {\footnotesize\textcolor{magenta}{(+ 0.41)}} & {\footnotesize\textcolor{magenta}{(+ 0.21)}}\tabularnewline
\midrule
\multirow{3}{*}{32} & \cellcolor{black!7}$E_{t}$ & 0 & 49.19 & 61.44 & 52.52 & 30.41 & 49.46 & 66.52\tabularnewline
 & $E_{b(32)}$ & $0.235$ & 49.57 & 61.92 & 52.92 & 30.68 & 49.75 & 66.73\tabularnewline
 &  &  & {\footnotesize\textcolor{magenta}{(+ 0.38)}} & {\footnotesize\textcolor{magenta}{(+ 0.48)}} & {\footnotesize\textcolor{magenta}{(+ 0.40)}} & {\footnotesize\textcolor{magenta}{(+ 0.27)}} & {\footnotesize\textcolor{magenta}{(+ 0.29)}} & {\footnotesize\textcolor{magenta}{(+ 0.21)}}\tabularnewline
\midrule
\multirow{3}{*}{64} & \cellcolor{black!7}$E_{t}$ & 0 & 45.84 & 57.54 & 48.97 & 27.73 & 46.16 & 63.22\tabularnewline
 & $E_{b(64)}$ & $0.174$ & 46.20 & 58.01 & 49.37 & 27.96 & 46.44 & 63.45\tabularnewline
 &  &  & {\footnotesize\textcolor{magenta}{(+ 0.36)}} & {\footnotesize\textcolor{magenta}{(+ 0.47)}} & {\footnotesize\textcolor{magenta}{(+ 0.40)}} & {\footnotesize\textcolor{magenta}{(+ 0.23)}} & {\footnotesize\textcolor{magenta}{(+ 0.28)}} & {\footnotesize\textcolor{magenta}{(+ 0.23)}}\tabularnewline
\midrule
\multirow{3}{*}{128} & \cellcolor{black!7}$E_{t}$ & 0 & 42.85 & 54.18 & 45.84 & 25.04 & 42.94 & 59.78\tabularnewline
 & $E_{b(128)}$ & $0.120$ & 43.14 & 54.55 & 46.16 & 25.20 & 43.15 & 59.99\tabularnewline
 &  &  & {\footnotesize\textcolor{magenta}{(+ 0.29)}} & {\footnotesize\textcolor{magenta}{(+ 0.37)}} & {\footnotesize\textcolor{magenta}{(+ 0.32)}} & {\footnotesize\textcolor{magenta}{(+ 0.16)}} & {\footnotesize\textcolor{magenta}{(+ 0.21)}} & {\footnotesize\textcolor{magenta}{(+ 0.21)}}\tabularnewline
\midrule
\multirow{3}{*}{256} & \cellcolor{black!7}$E_{t}$ & 0 & 40.42 & 51.48 & 43.33 & 22.57 & 39.96 & 56.83\tabularnewline
 & $E_{b(256)}$ & $0.078$ & 40.63 & 51.76 & 43.57 & 22.67 & 40.15 & 57.01\tabularnewline
 &  &  & {\footnotesize\textcolor{magenta}{(+ 0.21)}} & {\footnotesize\textcolor{magenta}{(+ 0.28)}} & {\footnotesize\textcolor{magenta}{(+ 0.24)}} & {\footnotesize\textcolor{magenta}{(+ 0.10)}} & {\footnotesize\textcolor{magenta}{(+ 0.19)}} & {\footnotesize\textcolor{magenta}{(+ 0.18)}}\tabularnewline
\bottomrule
\end{tabular}

}
\vspace{0.1em}

\caption{Subset Evaluation with Varying Subset Size. The CaliDet (DINO) is
evaluated on Objects365 subsets. Subset Evaluation with Varying Subset
Size. We split the whole validation dataset into subsets of varying
sizes. With a frozen model, we can still obtain some AP improvements
for free (\ie, without any gradient update) given an arbitrary distribution
shift in terms of the conditional probability.}

\label{tab:dino-mpio3}
\end{table}

\begin{figure}[h]
\includegraphics[width=\linewidth]{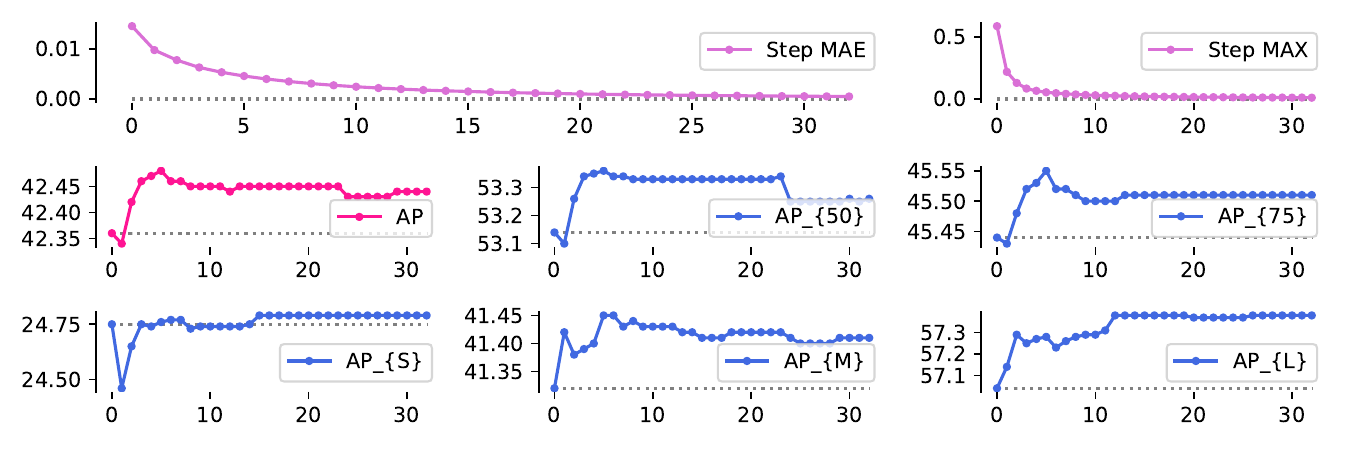}

\caption{Self-calibration results on O365 dataset using CaliDet (DINO) w/ subset
size 256.}

\label{fig:o3-256}
\end{figure}

\begin{figure}[h]
\includegraphics[width=\linewidth]{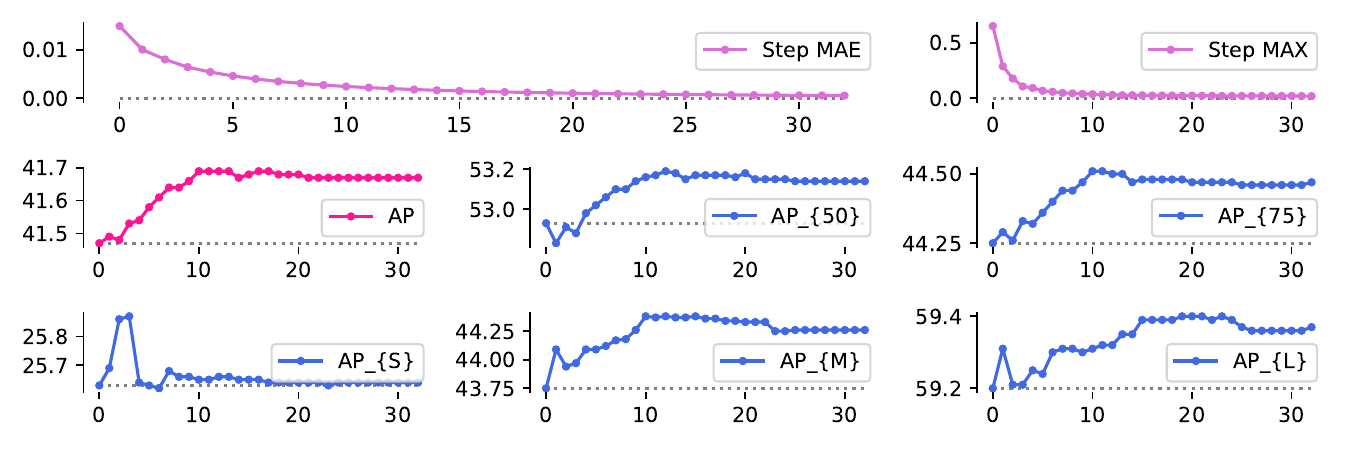}

\caption{Self-calibration results on O365 dataset using CaliDet (DINO) w/ subset
size 128.}

\label{fig:o3-128}
\end{figure}

\begin{figure}[h]
\includegraphics[width=\linewidth]{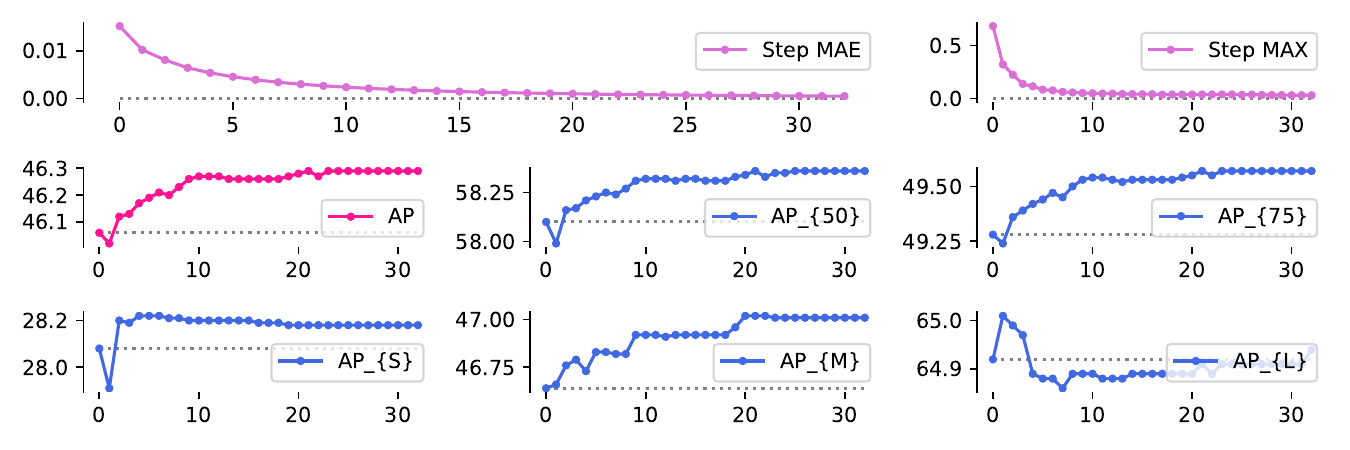}

\caption{Self-calibration results on O365 dataset using CaliDet (DINO) w/ subset
size 64.}

\label{fig:o3-64}
\end{figure}

\begin{figure}[h]
\includegraphics[width=\linewidth]{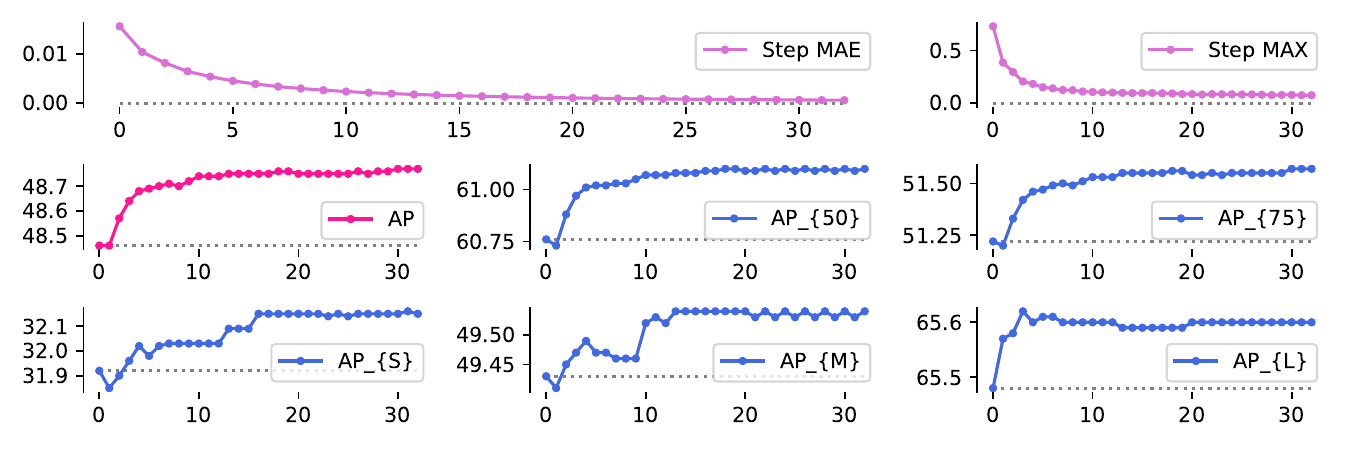}

\caption{Self-calibration results on O365 dataset using CaliDet (DINO) w/ subset
size 32.}

\label{fig:o3-32}
\end{figure}

\begin{figure}[h]
\includegraphics[width=\linewidth]{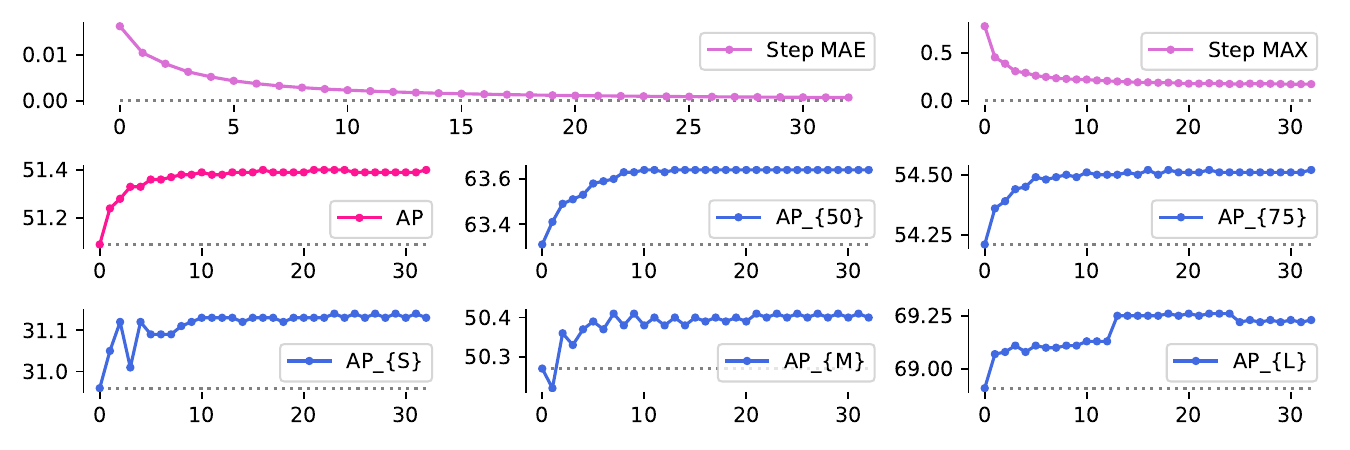}

\caption{Self-calibration results on O365 dataset using CaliDet (DINO) w/ subset
size 16.}

\label{fig:o3-16}
\end{figure}

\begin{figure}[h]
\includegraphics[width=\linewidth]{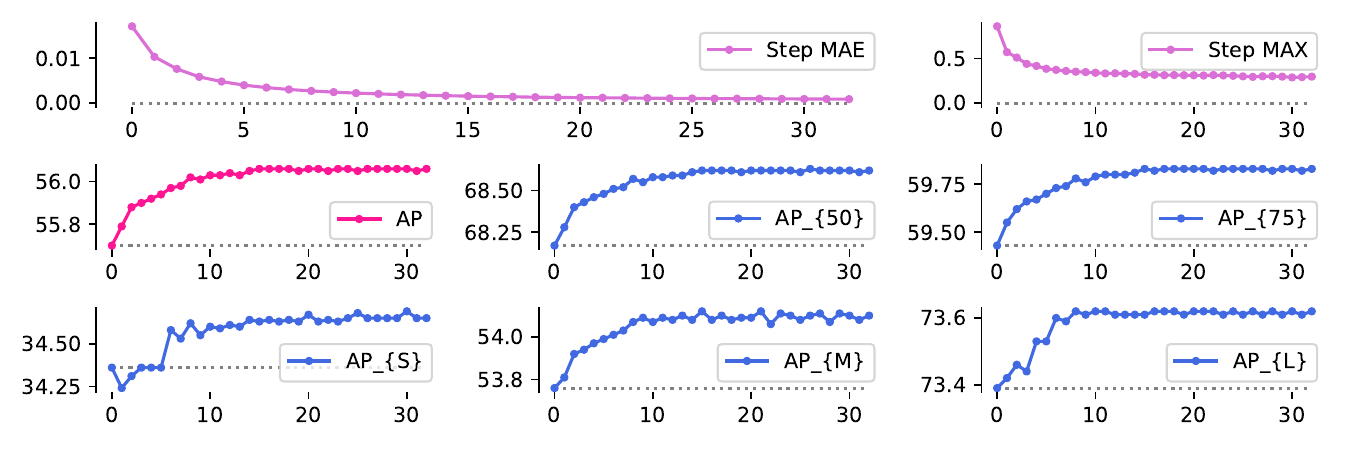}

\caption{Self-calibration results on O365 dataset using CaliDet (DINO) w/ subset
size 8.}

\label{fig:o3-8}
\end{figure}

We first evaluate the COCO model from the manuscript (\ie, based
on DINO). As shown in Tab.~\ref{tab:qc-1-o3}, our model is effective.
Specifically, although $E_{v}^{\text{O365}}$ deviates from $E_{t}^{\text{COCO}}$,
most of its conditional probabilities are still roughly correct. The
subset evaluation result can be found in Tab.~\ref{tab:dino-mpio3}.
The self-calibration results can be found in Fig.~\ref{fig:o3-256},
\ref{fig:o3-128}, \ref{fig:o3-64}, \ref{fig:o3-32}, \ref{fig:o3-16},
\ref{fig:o3-8}. 

The proposed CaliDet is effective, even if it is only trained on the
COCO training set (not seen any O365 sample during training). With
different deployment priors, the corresponding AP varies following
the expectation.

Apart from these, we are sorry for not having enough computational
resource to provide experimental results for the full Object365 dataset
(trained with all the 365 classes), or the LVIS dataset (with 1203
classes), because they require large-scale pre-training to reach a
sensible AP performance. During the training process, our method requires
a mostly correct matching in the decoder part to learn the correct
calibration vectors on the corresponding classification heads. If
the baseline model AP is low (for instance, an LVIS model without
large-scale pretraining), the calibration vector learning process
will be very noisy, and hence leads to a very weak calibration effect.
\end{document}